%% file: Stadion-arXiv.tex
\documentclass[runningheads]{llncs}
\usepackage{graphicx}
\usepackage{booktabs}
\usepackage{subfigure}
\usepackage{amsmath}
\usepackage{amsfonts}
\usepackage{url}
\usepackage{hyperref}
\usepackage{xcolor}
\usepackage{bbding}
\usepackage{enumitem}
\makeatletter
\newcommand{\printfnsymbol}[1]{%
  \textsuperscript{\@fnsymbol{#1}}%
}
\makeatother

\begin{document}
\title{Selecting the Number of Clusters $K$ with a Stability Trade-off: an Internal Validation Criterion}
%
\titlerunning{Selecting the Number of Clusters $K$ with a Stability Trade-off}
%
\author{Alex Mourer\thanks{Equal contribution. Supported by ANRT CIFRE grants and Safran Aircraft Engines.}\inst{1,4} \and
Florent Forest\printfnsymbol{1}\inst{2,4} \and 
Mustapha Lebbah\inst{3} \and 
Hanane Azzag\inst{2} \and
Jérôme Lacaille\inst{4}}  
%
\authorrunning{Mourer et al.}
%

\institute{SAMM, Université Paris 1 Panthéon Sorbonne, France\\
\and
LIPN (CNRS UMR 7030), Université Sorbonne Paris Nord, France\\
\email{\href{mailto:f@florentfo.rest}{f@florentfo.rest}}
\and
DAVID lab, Université de Versailles/Paris-Saclay, France\\
\and
Safran Aircraft Engines, France\\
}
%
\maketitle              
\begin{abstract}
Model selection is a major challenge in non-parametric clustering. There is no universally admitted way to evaluate clustering results for the obvious reason that no ground truth is available. The difficulty to find a universal evaluation criterion is a consequence of the ill-defined objective of clustering. In this perspective, clustering stability has emerged as a natural and model-agnostic principle: an algorithm should find stable structures in the data. If data sets are repeatedly sampled from the same underlying distribution, an algorithm should find similar partitions. However, stability alone is not well-suited to determine the number of clusters. For instance, it is unable to detect if the number of clusters is too small. We propose a new principle: a good clustering should be stable, and within each cluster, there should exist no stable partition. This principle leads to a novel clustering validation criterion based on between-cluster and within-cluster stability, overcoming limitations of previous stability-based methods. We empirically demonstrate the effectiveness of our criterion to select the number of clusters and compare it with existing methods. Code is available at \href{https://github.com/FlorentF9/skstab}{\url{https://github.com/FlorentF9/skstab}}.

\end{abstract}
%
%
%
\input{content/introduction}

\input{content/background}

\input{content/stability}

\input{content/stabinstab}
\input{content/experiments}

\input{content/conclusion}

\bibliography{references}
\bibliographystyle{splncs04}

\title{Selecting the Number of Clusters $K$ with a Stability Trade-off: an Internal Validation Criterion -- Supplementary Material}
%
\titlerunning{Supplementary Material}
%
\author{}
\authorrunning{Mourer et al.}
%
\institute{
}
\maketitle              
%
%
%
%
\input{supplementary/examples}
\input{supplementary/influence}
\input{supplementary/code}
\input{supplementary/experiment}

\bibliography{references}
\bibliographystyle{splncs04}

\end{document}

%% file: content/introduction.tex
\section{Introduction}\label{sec:introduction}

Clustering is an unsupervised learning technique aiming at discovering structure in unlabeled data. It can be defined as the “partitioning of data into groups (a.k.a. clusters) so that similar [...] elements share the same cluster and the members of each cluster are all similar” \cite{ben2018clustering}. These goals are contradictory because of the non-transitivity of similarity: if $A$ is similar to $B$, and $B$ is similar to $C$, $A$ is not necessarily similar to $C$. Since clustering is an ill-posed problem, it cannot be properly solved using this definition, and algorithms often optimize only one of its aspects. For instance, $K$-means only guarantees that dissimilar objects are separated, and on the other hand, single linkage clustering only guarantees that similar objects will end up in the same cluster. As a consequence, model selection is a major challenge in non-parametric clustering.

In the sample-based framework adopted in this work, model selection assesses whether partitions found by an algorithm correspond to meaningful structures of the underlying distribution, and not just artifacts of the algorithm or sampling process \cite{Smith1980,Shamir2007}. Practitioners need to evaluate clustering results in order to select the best parameters for an algorithm (e.g. the number of clusters $K$) or choose between different algorithms. Plenty of evaluation methods exist in literature, but they usually incorporate strong assumptions on the geometry of clusters or on the underlying distribution.

There is a need for a general, model-agnostic evaluation method. Clustering stability has emerged as a principle stating that "to be meaningful, a clustering must be both \emph{good} and the only \emph{good} clustering of the data, up to small perturbations. Such a clustering is called stable. Data that contains a stable clustering is said to be clusterable" \cite{meila2018tell}. Hence, a clustering algorithm should discover stable structures in the data. In statistical learning terms, if data sets are repeatedly sampled from the same underlying distribution, an algorithm should find similar partitions. As we do not have access to the data-generating distribution, perturbed data sets are obtained either by sampling or injecting noise into the original data. Stability seems to be an elegant principle, but there are still severe limitations in practice. For instance, stability does not necessarily depend on clustering outcomes but can be solely related to properties of the data such as symmetries \cite{ben2006sober}. As outlined in \cite{von2010clustering}, there exist various protocols to estimate stability. Unfortunately, a thorough study that evaluates them in practice is lacking.

\paragraph{Contributions} We propose a method for quantitatively and visually assessing the presence of structure in clustered data. The main contributions of our work can be stated as follows:
\begin{itemize}
    \item To our knowledge, this is the first large-scale empirical study on clustering stability analysis.
    \item A novel definition of clustering is proposed, based on between-cluster and within-cluster stability.
    Based on this definition, we introduce Stadion, the stability difference criterion, along with an interpretable visualization tool, called \emph{stability paths}.
    \item We show that additive noise perturbation is reliable, and a methodology to determine the amount of perturbation is proposed.
    \item We assess the ability of Stadion to select the number of clusters $K$ on a vast collection of data sets and compare it with state-of-the-art methods.
\end{itemize}

%% file: content/background.tex
\section{Related work}\label{sec:related}

Internal clustering indices measure the quality of a clustering when ground-truth labels are unavailable. Most criteria rely on a combination of between-cluster and within-cluster distances. Between-cluster distance measures how distinct clusters are dissimilar, while within-cluster distance measures how elements belonging to the same cluster are similar. Unfortunately, this incorporates a prior on the geometry of clusters \cite{Dunn1974,Calinski1974,Davies1979,Rousseeuw1987,Ray1999,Tibshirani2001,Desgraupes2013}.

Stability analysis for clustering validation is a long-established technique. It can be traced back as far as 1973 \cite{Strauss1973} and from there has drawn increasing attention \cite{Ben-Hur2002,Lange2004,ben2006sober,ben2007stability,Ben-David2008,von2010clustering}. Some works concluded that stability is not a well-suited tool for model selection \cite{Shamir2007}. In the general case, stability can only detect if the number of clusters is too large for the $K$-means algorithm (see Figure~\ref{fig:stab-example}). A partition with too few clusters is indeed stable, except for perfectly symmetric distributions. More accurately, these works proved that the asymptotic stability of risk-minimizing clustering algorithms, as sample size grows to infinity, only depends on whether the objective function has one or several global minima.

Albeit significant theoretical efforts, few empirical studies have been conducted. Each study focuses on specific practical implementations of stability, but as mentioned in \cite{von2010clustering,ben2014data}, a thorough study comparing all protocols in practice does not exist and a more objective evaluation of these results is warranted.

%% file: content/stability.tex
\section{Clustering stability}\label{sec:stability}

A data set $\mathbf{X} = \lbrace \mathbf{x}_1, \dots, \mathbf{x}_N \rbrace$ consists in $N$ independent and identically distributed (i.i.d.) samples, drawn from a data-generating distribution $\mathcal{P}$ on an underlying space $\mathcal{X}$. Formally, a clustering algorithm $\mathcal{A}$ takes as input the data set $\mathbf{X}$, some parameter $K \ge 1$, and outputs a clustering $\mathcal{C}_K = \lbrace  C_1, \ldots, C_K \rbrace$ of $\mathbf{X}$ into $K$ disjoint sets. Thus, a clustering can be represented by a function $\mathbf{X} \rightarrow \lbrace 1, \ldots, K \rbrace$ assigning a label to every point of the input data set. Some algorithms can be extended to construct a partition of the entire underlying space. This partition is represented by an extension operator function $\mathcal{X} \rightarrow \lbrace 1, \ldots, K \rbrace$ (e.g. for center-based algorithms, we compute the distance to the nearest center). 

\begin{figure*}[h]
    \centering
    \subfigure[$K = 2$ (stable)]{\includegraphics[width=0.32\linewidth]{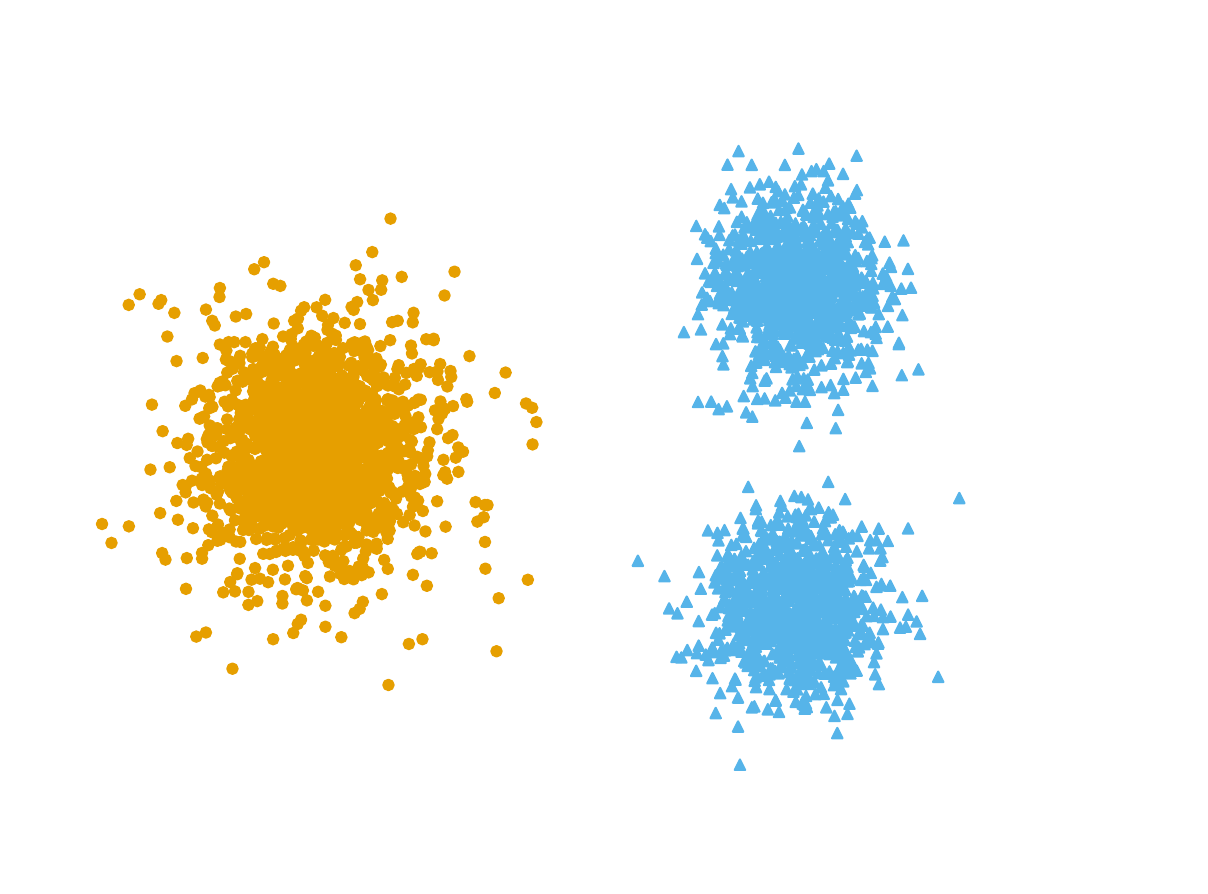}}
    \subfigure[$K = 3$ (stable)]{\includegraphics[width=0.32\linewidth]{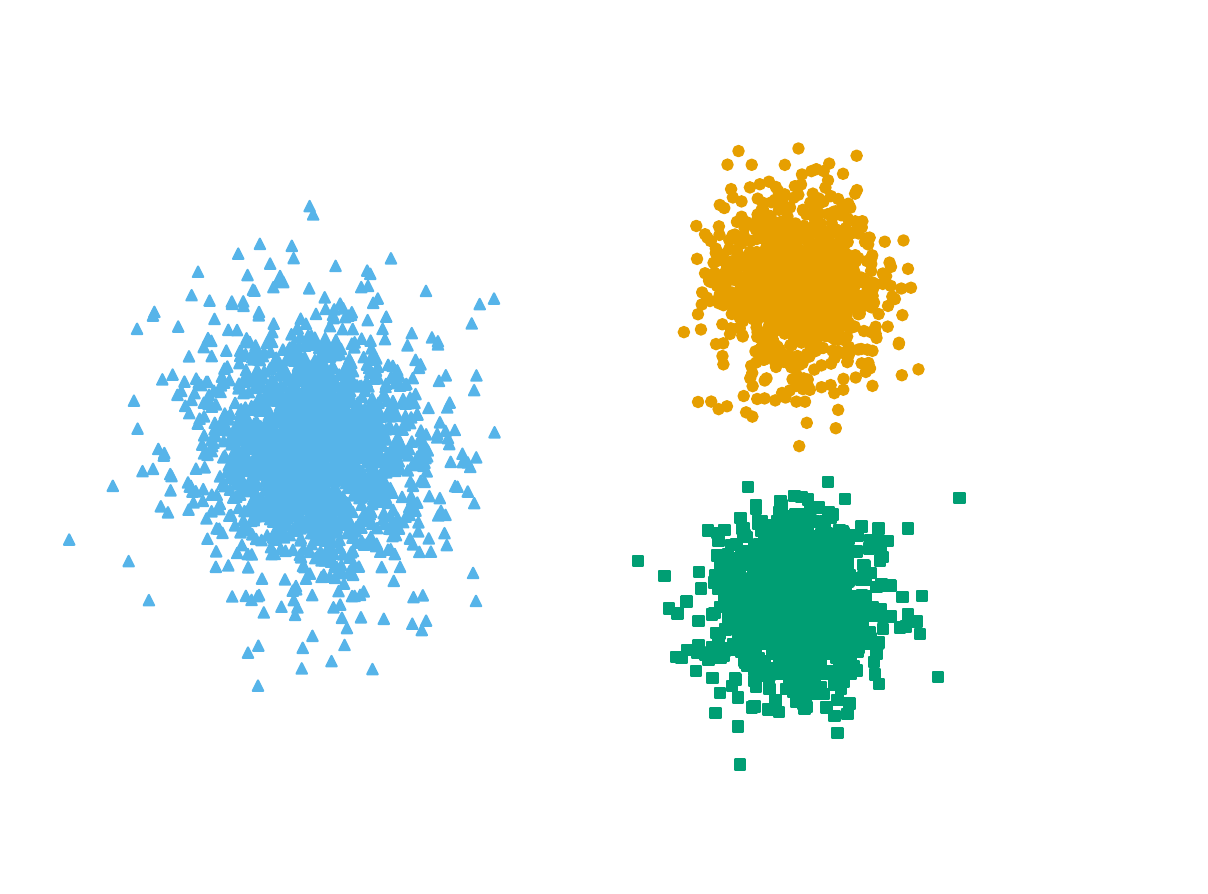}}
    \subfigure[$K = 4$ (unstable)]{\includegraphics[width=0.32\linewidth]{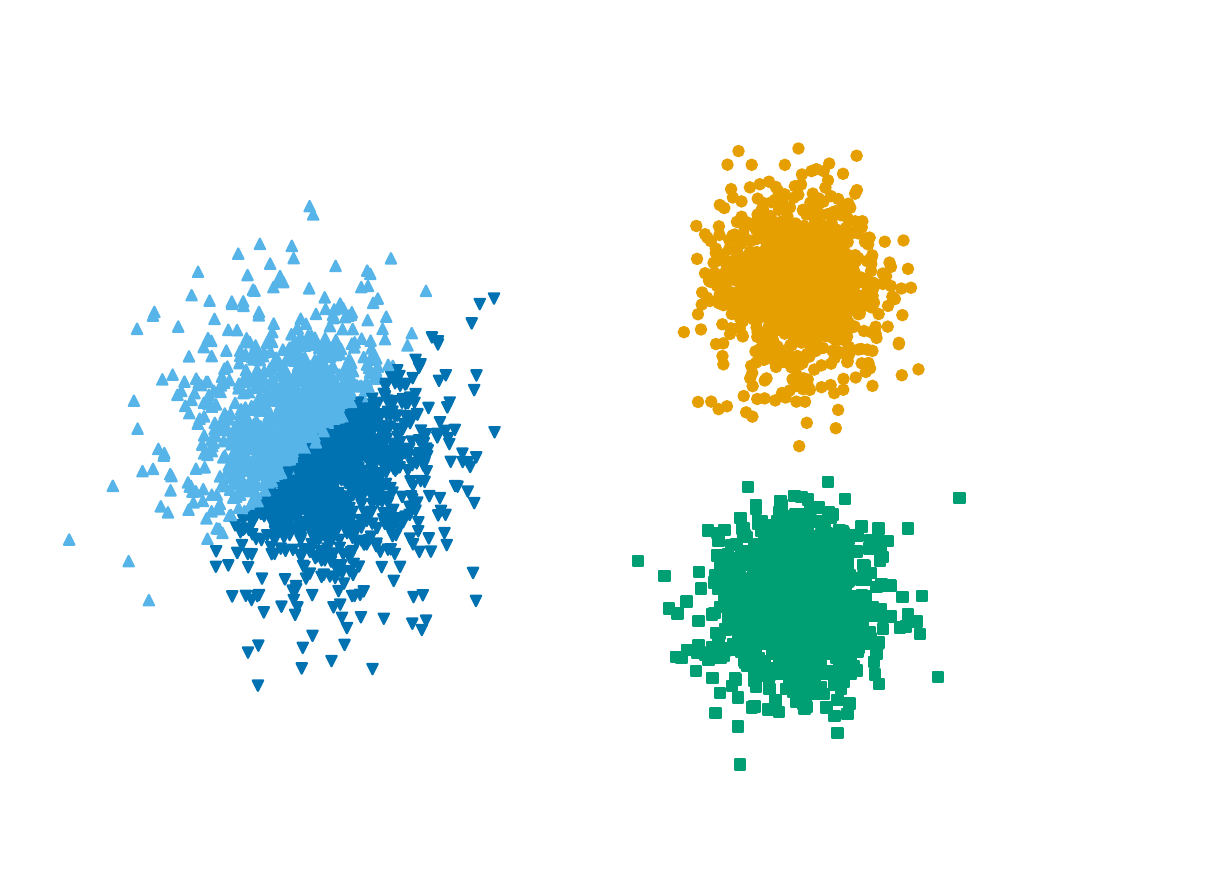}}
    \caption{Example data set with three clusters. The labels correspond to the $K$-means clustering result for $K = 2$, $3$ and $4$. $K$-means is stable even if $K$ is too small.}
    \label{fig:stab-example}
\end{figure*}

Let $\mathbf{X}$ and $\mathbf{X}'$ be two data sets drawn from the same distribution and note $\mathcal{C}_K$ and $\mathcal{C}_K'$ their respective clusterings. Let $s$ be a similarity measure such that $s(\mathcal{C}_K, \mathcal{C}_K')$ measures the agreement between the two clusterings. Then, for a given sample size $N$, the stability of a clustering algorithm $\mathcal{A}$ is defined as the expected similarity between $\mathcal{C}_K$ and $\mathcal{C}_K'$ on different data sets $\mathbf{X}$ and $\mathbf{X}'$, sampled from the same distribution $\mathcal{P}$,
\begin{equation}
    \text{Stab}(\mathcal{A}, K) := \mathbb{E}_{\mathbf{X}, \mathbf{X}' \sim \mathcal{P}^N} \left[ s(\mathcal{C}_K, \mathcal{C}_K') \right].
    \label{eq:stab}
\end{equation}
This quantity is unavailable in practice, as we have a finite number of samples, so it needs to be estimated empirically. Various methods have been devised to estimate stability using perturbed versions of $\mathbf{X}$.

The first methods used in literature are based on resampling the original data set (splitting in half \cite{Strauss1973}, subsampling \cite{Ben-Hur2002}, bootstrapping \cite{Falasconi2010,Fang2012}, jackknife \cite{Yeung2001}, etc.). Another method consists in adding random noise either to the data points \cite{Moller2006} or to their pairwise distances \cite{vijayaraghavan2017clustering,balcan2016clustering}. For high-dimensional data, alternatives are random projections or randomly adding or deleting variables \cite{Strauss1973}. Once the perturbed data sets are generated, there are several ways to compare the resulting clusterings. With noise-based methods, it is possible to compare the clustering of the original data set (reference clustering) with the clusterings obtained on perturbed data sets, or to compare only clusterings obtained on the latter. With sampling-based methods, we can compare overlapping subsamples on data points where both clusterings are defined \cite{Falasconi2010}, or compare clusterings of disjoint subsamples (using for instance an extension operator or a supervised classifier to transfer labels from one sample to another \cite{Lange2004}). Finally, possible similarity measures include external indices such as the ARI \cite{Falasconi2010,Zhao2011}.

Before discussing in details the  mechanisms of stability, we introduce a trivial example to illustrate its main issue: it cannot detect in general whenever $K$ is too small.
Consider the example presented in Figure~\ref{fig:stab-example} with three clusters. On any sample from such a distribution, as soon as we have a reasonable amount of data, $K$-means with $K = 2$ always constructs the solution separating the left cluster from the two right clusters. Consequently, it is stable despite $K = 2$ being the wrong number of clusters. This situation was pointed out in \cite{ben2006sober}.
\begin{figure*}[h]
    \centering
    \centerline{\includegraphics[width=\textwidth]{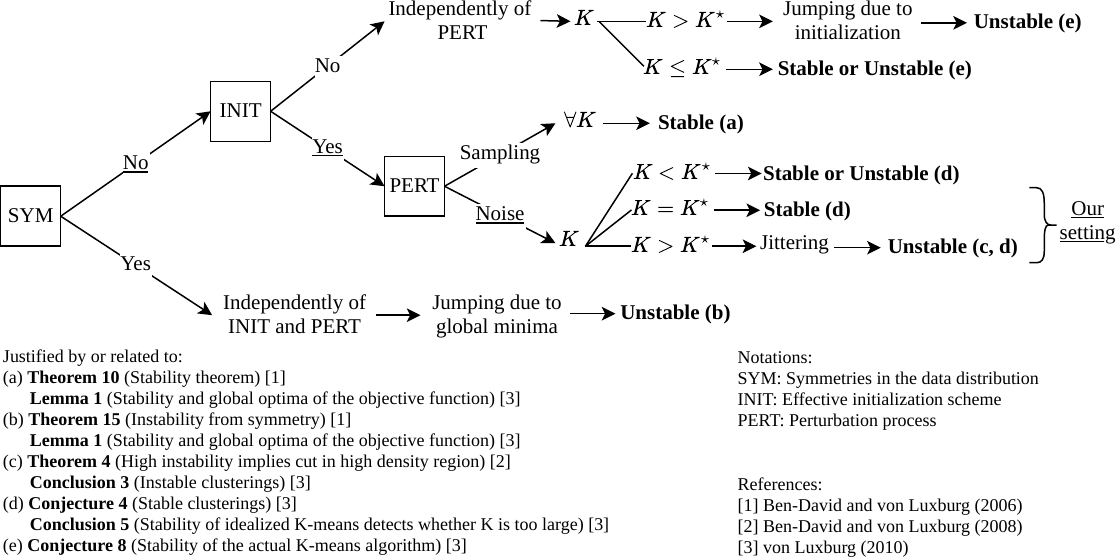}}
    \caption{Diagram explaining sources of instability in different settings, based on theoretical results for $K$-means, with large sample size, assuming $K \ll N$ and the underlying distribution has $K^{\star}$ well-separated clusters that can be represented by $K$-means. We consider no symmetries, effective initialization and noise-based perturbation, thus instability (due to jittering) arises when $K$ is too large, and when $K$ is too small whenever cluster boundaries are in high-density regions.}
    \label{fig:stab-diagram}
\end{figure*}

In the case of algorithms that minimize an objective function (e.g. center-based or spectral), two different sources of instability have been identified \cite{von2010clustering}. First, \emph{jittering} is caused by assignment changes at cluster boundaries after perturbation. Therefore, strong jitter is produced when a cluster boundary cuts through high-density regions. Second, \emph{jumping} refers to the algorithm ending up in different local minima. The most important cause of jumping is initialization. Another cause is the presence of several global minima of the objective function. This happens if there are perfect symmetries in the distribution, which is very unlikely in real-world data. Examples are provided as supplementary material.



However, practitioners mainly use algorithms with consistent initialization strategies. For instance with $K$-means, we keep the best trial over a large number of runs and use the $K$-means++ heuristic. This initialization tends to make $K$-means deterministic, differently from the random initialization proposed in \cite{von2010clustering,Bubeck2012}, which allows jumping to occur whenever $K > K^{\star}$, where $K^{\star}$ is the true number of clusters. 
Throughout this work, we consider a setting with large enough sample size, without perfect symmetries and with consistent initialization, that we deem to be realistic. Thus, we do not rely on jumping as the main source of instability even when $K > K^{\star}$, and rather rely on jittering. 
As a consequence, we need a perturbation process that produces jittering. We settle for noise-based perturbation, because as soon as $N$ is reasonably large, resampling methods become trivially stable whenever there is a single global minimum \cite{ben2006sober,von2010clustering}. We summarize important results in the diagram Figure~\ref{fig:stab-diagram} and provide a simple example where sampling methods such as \cite{Ben-Hur2002,Lange2004} fail in the supplementary material.
To conclude, in our setting, a noise-based perturbation process causes jittering, enabling stability to indicate whenever $K$ is too large. On the other hand, stability cannot in general detect when $K$ is too small.
In order to overcome this limitation, we introduce the concept of \textit{within-cluster stability}. 

%% file: content/stabinstab.tex
\section{Between-cluster and within-cluster stability}\label{sec:stabinstab}

A clustering algorithm applied with the same parameters to perturbed versions of a data set should find the same structure and obtain similar results. The stability principle described by (\ref{eq:stab}) relies on between-cluster boundaries and we thus call it \textit{between-cluster stability}. Therefore, it cannot detect structure within clusters. In Figure~\ref{fig:stab-example}, $K=2$ is stable, whereas one cluster contains two sub-clusters. This sub-structure cannot be detected by between-cluster stability alone. Obviously, this implies that stability is unable to decide whether a data set is clusterable or not (i.e. when $K^{\star} = 1$), which is a severe limitation. For this very reason, we introduce a second principle of \textit{within-cluster stability}: clusters should not be composed of several sub-clusters. This implies the absence of stable structures inside any cluster. In other words, any partition of a cluster should be unstable. The combination of these two principles leads to a new definition of a clustering:
\begin{definition}{}
 A clustering is a partitioning of data into groups so that the partition is stable, and within each cluster, there exists no stable partition. 
\end{definition}
A clustering should have a high between-cluster stability and a low within-cluster stability. Despite its apparent simplicity, implementing this principle is a difficult task. As seen in the last section, between-cluster stability can be estimated in many ways. On the other hand, within-cluster stability is a challenging quantity to define and estimate.  
We propose a method to estimate both quantities, and then we detail and discuss our choices.

%
\subsection{Stadion: a novel stability-based validity index}

Let $\{\mathbf{X}_{1} , \ldots, \mathbf{X}_{D}\}$ be $D$ perturbed versions of the data set obtained by adding random noise to the original data set $\mathbf{X}$. Between-cluster stability of algorithm $\mathcal{A}$ with parameter $K$ estimates the expectation (\ref{eq:stab}) by the empirical mean of the similarities $s$ between the reference clustering $\mathcal{C}_K = \mathcal{A}(\mathbf{X}, K)$ and the clusterings of the perturbed data sets,
\begin{align}
    \text{Stab}_{\text{B}}(\mathcal{A}, \mathbf{X}, K) := \frac{1}{D} \sum_{d = 1}^D s\left(\mathcal{A}(\mathbf{X}, K), \mathcal{A}(\mathbf{X}_{d}, K)\right).
    \label{eq:stabb}
\end{align}
Since $s$ is a similarity measure, this quantity needs to be maximized. In order to define within-cluster stability, we need to assess the presence of stable structures inside each cluster. To this aim, we propose to \textit{cluster again} the data within each cluster of $\mathcal{C}_K$. Formally, let $\mathbf{\Omega} \subset \mathbb{N}^{*}$ be a set of numbers of clusters. The $k$-th cluster in the reference clustering is noted $C_k$, its number of elements $N_k$. Within-cluster stability of algorithm $\mathcal{A}$ is defined as
\begin{align}
    \text{Stab}_{\text{W}}(\mathcal{A}, \mathbf{X}, K, \mathbf{\Omega}) :=
    \sum_{k = 1}^K  \bigg( \frac{1}{| \mathbf{\Omega} | } \sum_{K' \in \mathbf{\Omega}}  \text{Stab}_{\text{B}}(\mathcal{A}, C_k, K')  \bigg) \times \frac{N_k}{N}.
    \label{eq:stabw}
\end{align}
As a good clustering is unstable within each cluster, this quantity needs to be minimized. Hence, we propose to build a new validity index combining between-cluster and within-cluster stability. A natural choice is to maximize the difference between both quantities. We call this index \textit{Stadion}, standing for \textit{stability difference criterion}:
\begin{equation}
    \text{Stadion}(\mathcal{A}, \mathbf{X}, K, \mathbf{\Omega}) := \text{Stab}_{\text{B}}(\mathcal{A}, \mathbf{X}, K) -\text{Stab}_{\text{W}}(\mathcal{A}, \mathbf{X}, K, \mathbf{\Omega}).
    \label{eq:stadion}
\end{equation}
The same partition $\mathcal{C}_K = \mathcal{A}(\mathbf{X}, K)$ is used in both terms of (\ref{eq:stadion}). Thus, Stadion evaluates the stability of an algorithm w.r.t. a reference partition.
%
\paragraph{How to perturb data?} We consider the setting in Figure~\ref{fig:stab-diagram} that is deemed to be realistic. 
Neither jumping nor jittering will occur if data are perturbed by sampling processes, as soon as there is enough data. Therefore, only noise-based perturbation is considered here. Among them, we adopt the $\varepsilon$-Additive Perturbation ($\varepsilon$-AP) with Gaussian or uniform noise, assuming variables are scaled to zero mean and unit variance. The number of perturbations $D$ can be kept very low and still gives reliable estimates (an analysis on the influence of $D$ is conducted in the supplementary material. 
\paragraph{How to choose $\varepsilon$?} A central trade-off has to be taken into account when perturbing the data set. If the noise level $\varepsilon$ is too strong, we might alter the very structure of the data.
We propose to circumvent this issue by \emph{not} choosing a single value for $\varepsilon$, but a grid of values. By gradually increasing $\varepsilon$ from $0$ to a value $\varepsilon_{\text{max}}$, we obtain what we call a \textit{stability path}, i.e. the evolution of stability as a function of $\varepsilon$. This method has one crucial advantage: it allows to compare partitions for different values of $\varepsilon$ without the necessity of choosing one. However, it comes with two drawbacks: setting both the fineness and the maximum value of the grid. In our experiments, the fineness does not play a major role in the results. A straightforward method to fix a maximum value $\varepsilon_{\text{max}}$ beyond which comparisons are not meaningful anymore is as follows. The perturbation corresponding to $\varepsilon_{\text{max}}$ is meant to destroy the cluster structure of the original data. This corresponds to the value where the data are no longer clusterable, i.e. $K = 1$ becomes the solution with the best Stadion value. A first guess at $\varepsilon_{\text{max}} = \sqrt{p}$ (where $p$ is the data dimension) works well in practice. We found that visualizing the stability paths (see Figure~\ref{fig:stab-paths}) greatly helps interpreting the structures found by an algorithm, hence improving the \textit{usefulness} of results.
\paragraph{How to compare partitions?} The similarity measure $s$ chosen to compare two partitions is the ARI. Note that it is used to compare cluster assignments and not the ground-truth labels. Its value is in $[0, 1]$, thus Stadion has a value in $[-1, 1]$, with $1$ corresponding to the best clustering and $-1$ to the worst. A total of 16 different similarity measures (such as the NMI) were compared (results of this study are in supplementary material).
\paragraph{How to aggregate the Stadion path?} To compute a scalar validity index for model selection, the Stadion path must be aggregated on the noise strength $\varepsilon$ from $0$ to $\varepsilon_{\text{max}}$. Two aggregation strategies, the maximum (Stadion-max) and the mean (Stadion-mean), are evaluated in our experiments.

The within-cluster stability is governed by the parameter $\mathbf{\Omega}$, which detects stable structures inside clusters of $\mathcal{C}_K$. As these are unknown, averaging several different values in $\mathbf{\Omega}$ gives a better estimate. In absence of sub-clusters, all partitions will be unstable because cluster boundaries will be placed in high-density regions. For the opposite reason, in presence of sub-clusters, at least some partitions will result in higher stability, thus increasing the within-cluster stability.
The analysis of influence conducted in supplementary material shows that $\mathbf{\Omega}$ has low impact on Stadion results and can be set easily.


An important assumption behind our implementation of within-cluster stability is that for non-clusterable structures (e.g. uniform noise), the algorithm will place cluster boundaries in high-density regions to produce instability through jittering. This encompasses a wide range of algorithms such as center-based, spectral or Ward linkage clustering which, for the sake of saving cost, would cut through dense clouds of points. If this requirement is not fulfilled, further studies are needed to determine whether this method will work.

%% file: content/experiments.tex
\section{Experiments}


%
\subsection{A simple example with stability paths}

\begin{figure*}[ht]
    \centering
    \includegraphics[width=0.45\textwidth]{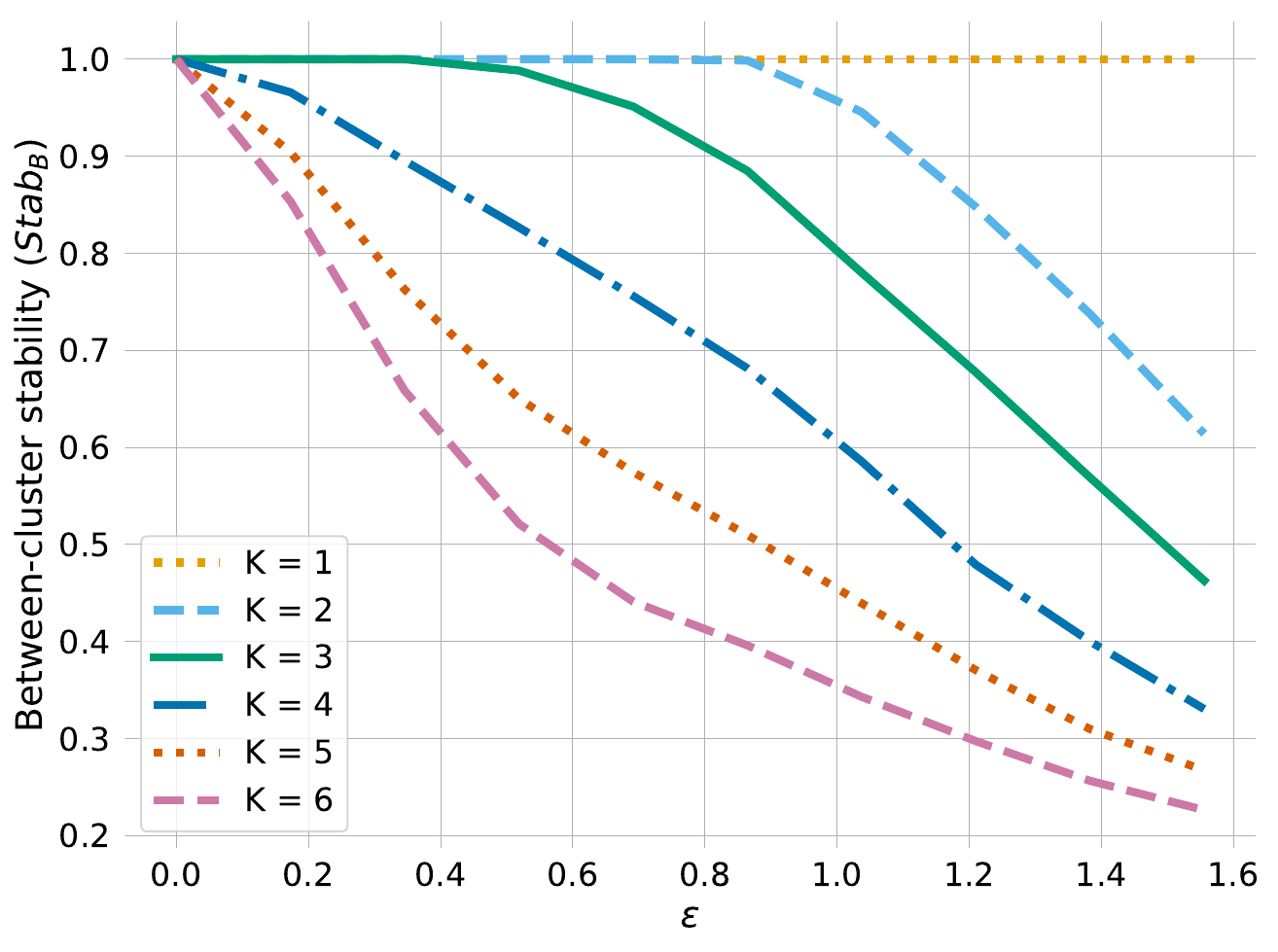}
    \includegraphics[width=0.45\textwidth]{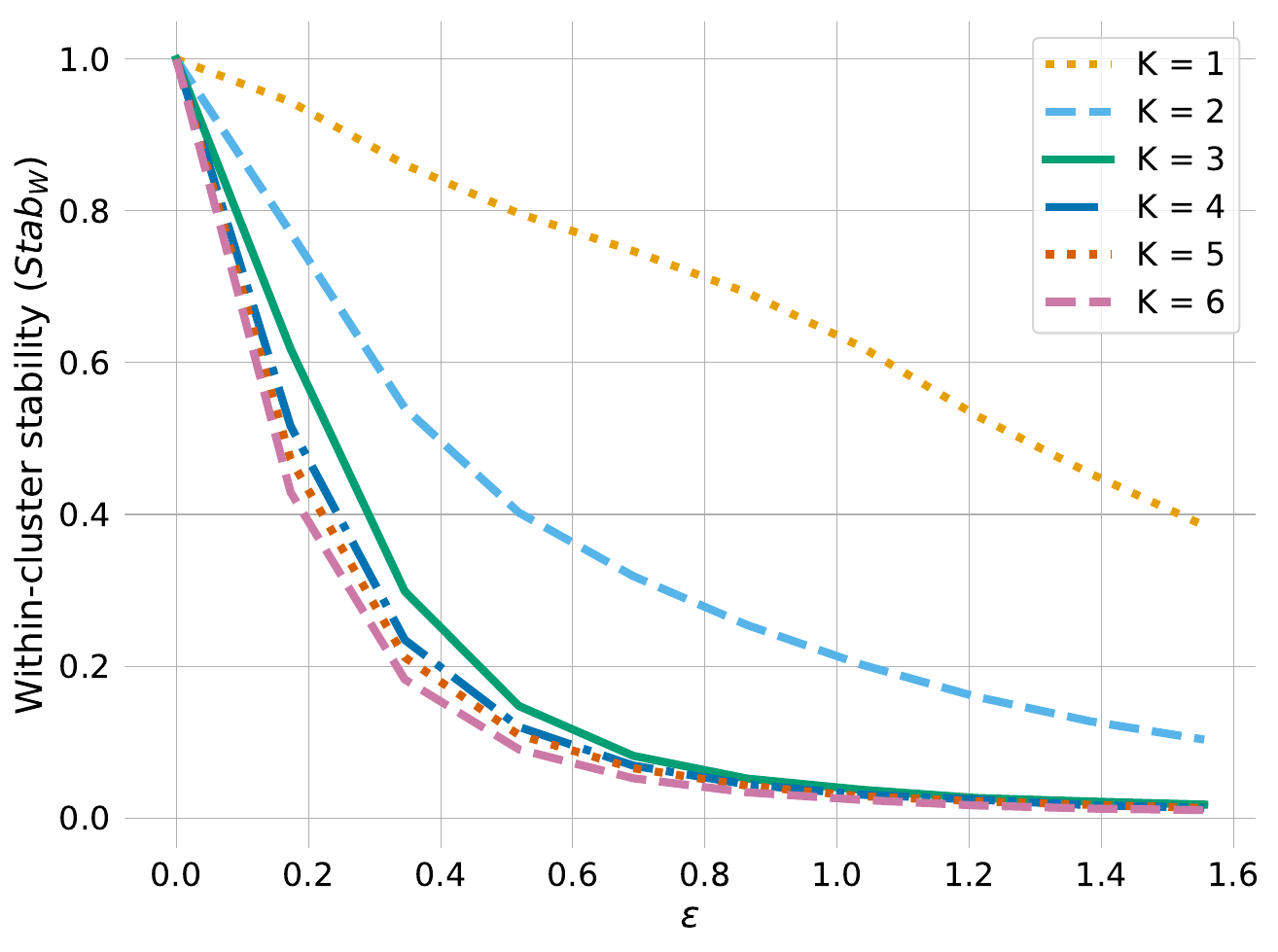}\\
    \includegraphics[width=0.45\textwidth]{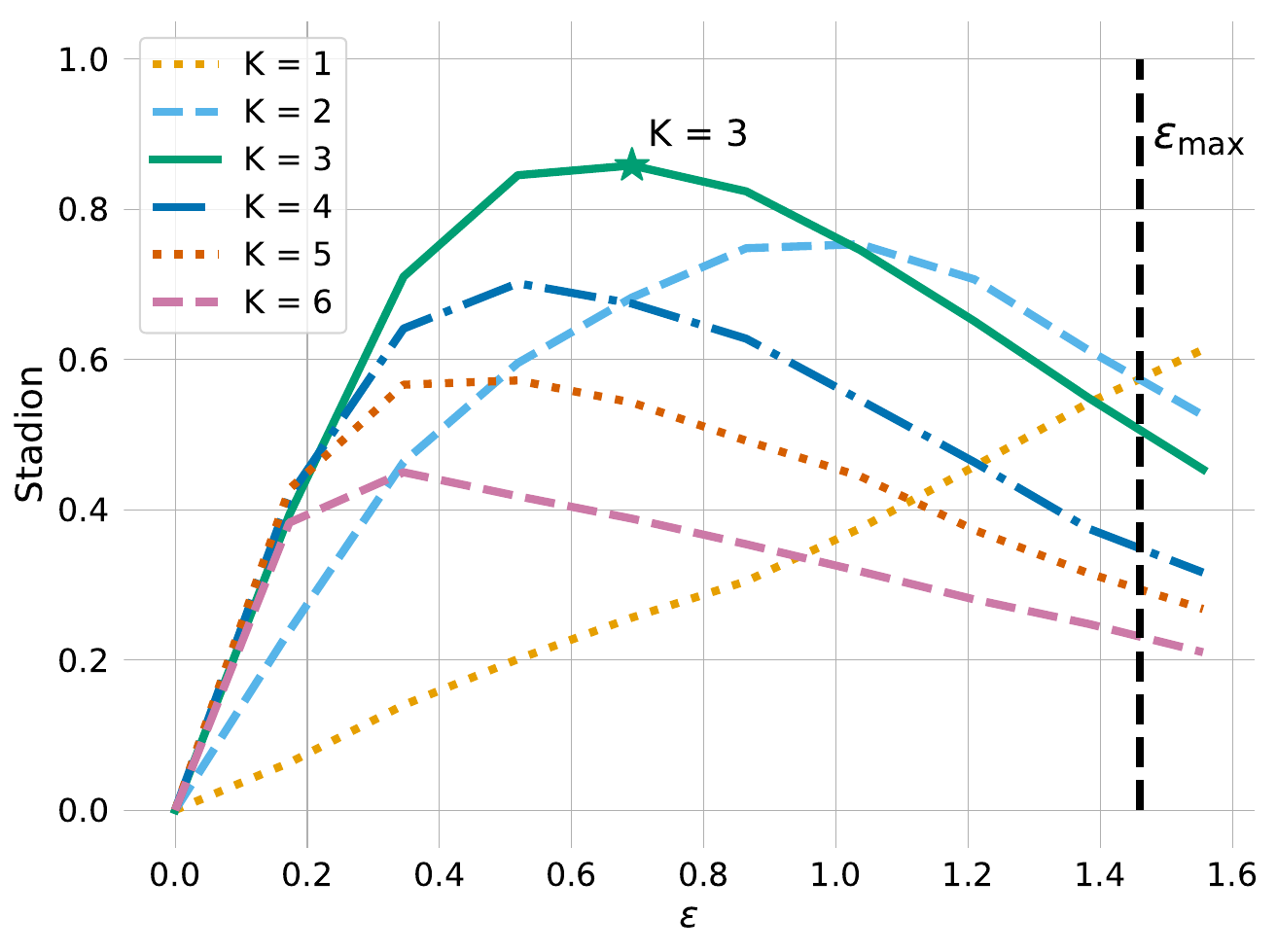}
    \includegraphics[width=0.45\textwidth]{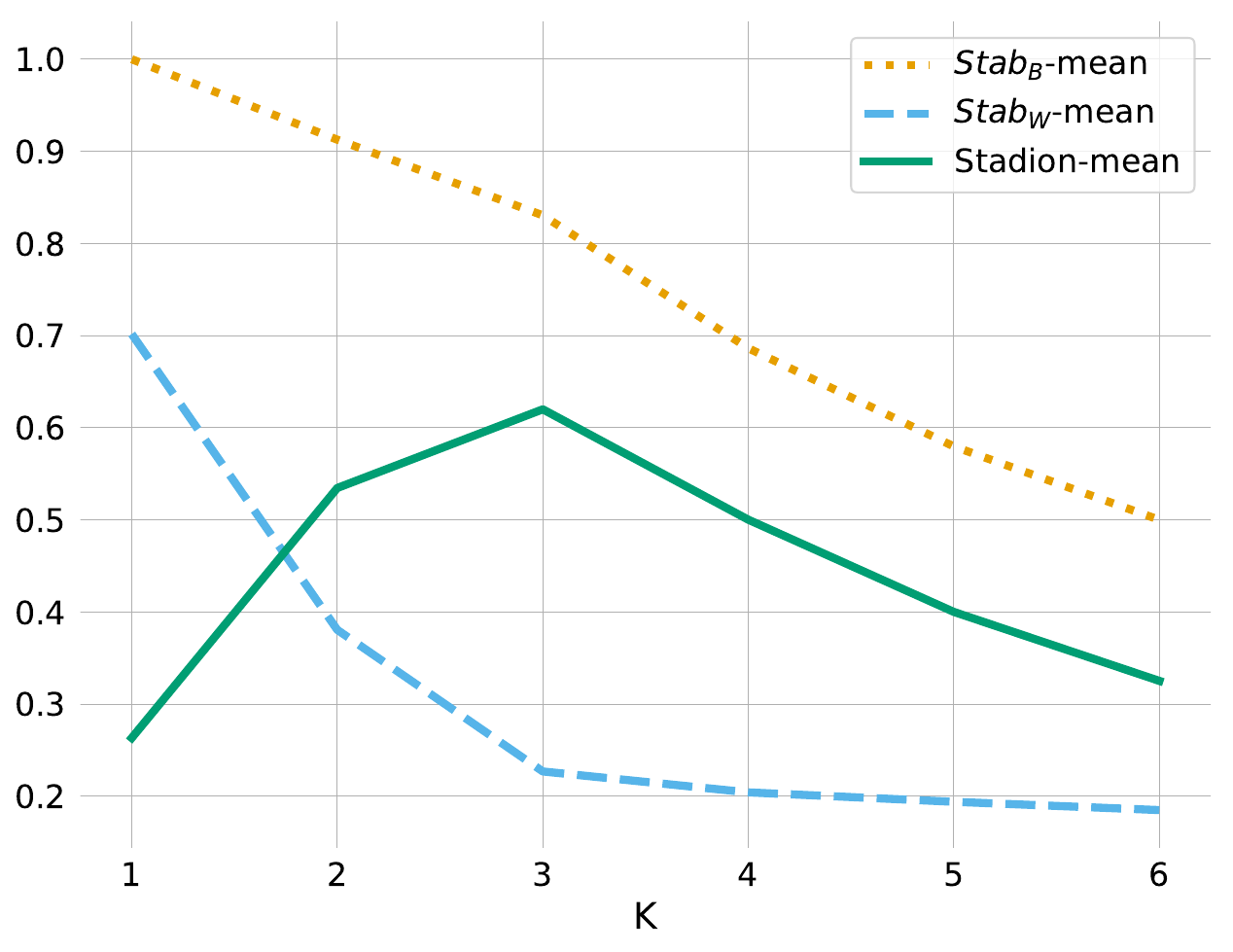}
    \caption{Between-cluster stability paths (top left), within-cluster stability paths (top right), Stadion paths (bottom left) and stability trade-off curve (bottom right) for $K$-means on the data set shown in Figure~\ref{fig:stab-example}, for $K \in \{1 \ldots 6\}$. $\varepsilon$ is the amplitude of the uniform noise perturbation. The best solution $K=3$ is selected either by taking the maximum or by averaging Stadion over $\varepsilon$ until $\varepsilon_{\text{max}}$. The trade-off plot represents the averaged Stadion, between- and within-cluster stability as a function of $K$.}
    \label{fig:stab-paths}
\end{figure*}

We begin by illustrating our method with $K$-means and uniform $\varepsilon$-AP on the example discussed previously (see Figure~\ref{fig:stab-example}). Figure~\ref{fig:stab-paths} displays between-cluster stability, within-cluster stability and Stadion as a function of the noise strength $\varepsilon$. For reasonable amounts of noise, the solutions $K=1$, $K=2$ and $K=3$ are all perfectly stable, showing the insufficiency of between-cluster stability alone to indicate whenever $K$ is too small. The solutions for $K \geq 4$ cut through the clusters and are thus unstable due to jittering. However, the solutions for $K=1$ and $K=2$ both have high within-cluster stability, caused by the presence of sub-clusters, which is not the case for $K \geq 3$. By computing a difference, our criterion Stadion combines this information and is able to indicate the correct number of clusters ($K=3$) by selecting the Stadion path with the highest maximum or mean value. Through its formulation, Stadion is acting as a \emph{stability trade-off}. The stability paths also give additional insights about the data structure. For example, we can read from the between-cluster stability path how the clusters successively merge together as $\varepsilon$ increases. Finally, the last graph (called stability trade-off plot) represents Stadion-mean 
for different values of $K$.

\subsection{Benchmark of clustering validation methods}

\paragraph{Methodology.} Importantly, we aim at evaluating clustering validation methods and not the clustering algorithms themselves. Thus, we evaluate methods on a large collection of 73 artificial benchmark data sets, most of them extensively used in literature, with a guaranteed known ground-truth cluster structure. Most data sets are available in \cite{deric,gago} and all data will be shared after publication. It was ensured that the evaluated algorithms are able to obtain good solutions (i.e., reasonably close to the ground-truth clustering), for some optimal parameter $K$. The data sets also provide various difficulty levels by varying the numbers, sizes, variances, shapes of clusters and noisy, close-by or overlapping clusters.

To compare the different validation methods, we first report the number of data sets where each method found $K^{\star}$, which we refer to as the number of \emph{wins}.
However, only checking whether $K^{\star}$ is selected is not always related to the goodness of the partition w.r.t. ground-truth, as the algorithm does not necessarily succeed in finding a good partition into $K^{\star}$ clusters.

Thus, we also compute the ARI between the selected partition and the ground-truth. Let us note $\mathcal{Y}_{K^{\star}} = \lbrace  Y_1, \ldots, Y_{K^{\star}} \rbrace$ the ground-truth partition. The performance of each method is assessed by computing $\text{ARI}(\mathcal{Y}_{K^{\star}}, \mathcal{C}_{\hat{K} } )$, where $\hat{K}$ is the estimated number of clusters. In order to compare methods over multiple data sets, we compute the average ranks, denoted $\overline{R_{\text{ARI}}}$. Since data sets have different difficulties, their results are not comparable and simply reporting an average would be meaningless \cite{demvsar2006statistical}. Thus, comparing their ranks is a more sound and fair solution.

%
%

\begin{table}[t]
  \caption{Benchmark results on 73 data sets for $K$-means, Ward and GMM. Average rank of the ARI between the selected clustering and ground-truth clustering ($\overline{R_{\text{ARI}}}$) and number of times the ground-truth $K^{\star}$ was selected (wins). 
  }
  \label{tab:all-benchmark}
  \centering
    \small{
    \begingroup
    \setlength{\tabcolsep}{3.8pt} 
    \begin{tabular}{lcccccc}
    \toprule
    & \multicolumn{2}{c}{$K$-means} & \multicolumn{2}{c}{Ward} & \multicolumn{2}{c}{GMM} \\
    Method & $\overline{R_{\text{ARI}}}$ & wins & $\overline{R_{\text{ARI}}}$ & wins & $\overline{R_{\text{ARI}}}$ & wins \\
    \midrule
    $K^{\star}$ (Oracle) & 8.11 & 73 & 4.77 & 73 & 5.05 & 73  \\
    \midrule
    Stadion-max & \textbf{7.46} & 50 & \textbf{5.25} & \textbf{54} & - & - \\
    Stadion-mean & 7.70 & 51 & 5.80 & 49 & - & -  \\
    Stadion-max (extended) & 7.58 & \textbf{56} & - & - & \textbf{5.59} & \textbf{56} \\
    Stadion-mean (extended) & 8.09 & 48 & - & - & 6.79 & 43 \\
    BIC & - & - & - & - & 6.45 & 48\\
    Wemmert-Gancarski \cite{Desgraupes2013} & 8.33 & 53 & 5.40 & \textbf{54} & 5.77 & 52 \\
    Silhouette \cite{Rousseeuw1987} & 9.55 & 46 & 6.47 & 45 & 7.01 & 45  \\
    Lange \cite{Lange2004} & 10.18 & 45 & 6.53 & 51 & 6.99 & 48  \\
    Davies-Bouldin \cite{Davies1979} & 10.21 & 40 & 6.45 & 41 & 7.29 & 34 \\
    Ray-Turi \cite{Ray1999} & 10.28 & 37 & 6.97 & 40 & 7.68 & 33 \\
    Hennig \cite{hennig2007cluster} & 10.72 & 37 & - & - & - & - \\
    Calinski-Harabasz \cite{Calinski1974} & 11.44 & 41 & 7.14 & 39 & 7.43 & 37 \\
    Gap statistic (B) \cite{Tibshirani2001} & 11.49 & 29 & - & - & - & - \\
    X-means \cite{Pelleg2000} & 11.56 & 28 & - & - & - & - \\
    Dunn \cite{Dunn1974} & 13.09 & 26 & 7.77 & 33 & 7.92 & 34 \\
    Hofmeyr \cite{hofmeyr2018degrees} & 13.20 & 30 & - & - & - & - \\
    Xie-Beni & 13.30 & 22 & 7.61 & 34 & 8.19 & 28 \\ 
    Gap statistic (A) \cite{Tibshirani2001} & 13.57 & 26 & - & - & - & - \\
    G-means \cite{Hamerly2004} & 13.74 & 24 & - & - & - & - \\
    Ben-Hur \cite{Ben-Hur2002} & 14.34 & 20 & 7.86 & 31 & 8.85 & 28 \\
    SpecialK \cite{Hess2019} & 17.07 & 19 & - & - & - & - \\
    \bottomrule
    \end{tabular}
    \endgroup
    }
\end{table}

In this benchmark, three algorithms are considered: $K$-means, Gaussian Mixture Models (GMM) and Ward hierarchical clustering. For $K$-means, two versions of Stadion are evaluated: the first one using the stability computation described in Section 4.1 (referred to as the \emph{standard} version), and the second one with an approximation using the extension operator (referred to as the \emph{extended} version). As seen in Section 3, an extension operator extends a clustering to new data points. $K$-means extends naturally by computing the Euclidean distance to centers. Hence, instead of re-running $K$-means for each perturbation, we directly predict the cluster assignments of perturbed data points. 
GMM allows a similar extension, by assigning points to the cluster with the highest posterior probability. It is the only version considered due to GMM's computational cost.

Table~\ref{tab:all-benchmark} summarizes results for each algorithm and validation method. We evaluated $K \in \{1, \ldots, 60\}$. For Stadion, we used uniform noise, $D=10$, $\mathbf{\Omega} = \{ 2,\ldots,10 \}$ and $s=\text{ARI}$. We also evaluate the partitions obtained with the ground-truth $K^{\star}$ and a selection of widely used clustering validation indices (when applicable) \cite{Desgraupes2013}, the Gap statistic \cite{Tibshirani2001} (with alternative versions A and B implemented in \cite{maechler2013package}), BIC, X-means \cite{Pelleg2000}, G-means \cite{Hamerly2004}, the Hennig procedure \cite{hennig2007cluster}, stability methods \cite{Ben-Hur2002,Lange2004}, and the recent SpecialK \cite{Hess2019}. For SpecialK, we used the default parameters indicated by the authors, but the assumptions made by the method failed on 11 data sets, explaining the poor results. Unfortunately, other methods like dip-means or skinny-dip did not have easy-to-use available implementations and were not included in this study. 

Stadion-max achieves the best results overall. On $K$-means, it is even ranked higher than the Oracle in terms of ARI. The second-best performing index is Wemmert-Gancarski (WG). It was shown in \cite{balcan2016clustering} that agglomerative clustering is not robust to noise, which explains inferior Stadion results with Ward. Moreover, results are slightly biased in favor of the indices that are only valid for $K \geq 2$, unlike Stadion that will select $K=1$ on non-clusterable distributions. 

%% file: content/conclusion.tex
\section{Conclusion}

In this paper, we tackled some of limitations of cluster stability for model selection. Our contribution is twofold. First, stability can be well estimated through additive noise perturbation, solving the limitations of sampling-based stability methods. Second, we introduce the Stadion (stability difference criterion), a novel criterion acting as a trade-off between traditional stability and within-cluster stability.
Furthermore, our method to control the amount of perturbation provides an interpretable visualization called stability paths.

We evaluated Stadion and methods of the literature on 73 clustering benchmark data sets. 
Performance is superior or on par with internal clustering indices that were designed with specific cluster geometries in mind, while relying on more general assumptions. This comes at a computational cost, requiring to run the algorithm many times. Nevertheless, studies have shown that it can be drastically reduced by down-sizing the hyperparameters with negligible impact on performance.
Moreover, most theoretical results used here were derived for $K$-means, and more work is needed to extend these concepts to other algorithms.

Altogether, model selection remains a challenge and there is yet no theory nor a methodology that can fulfill this task perfectly. We proposed an empirical method showing interesting results along with hints to a theoretical background that could be established in future work. We hope that it will spark much-needed interest in the research community to further advance this field.


%% file: supplementary/examples.tex
\section{Additional experiments and examples}\label{appendix:examples}

%
\subsection{Finding $K=1$: the case of non-clusterable data}

\begin{figure}[h]
    \centering
    \includegraphics[width=0.49\columnwidth]{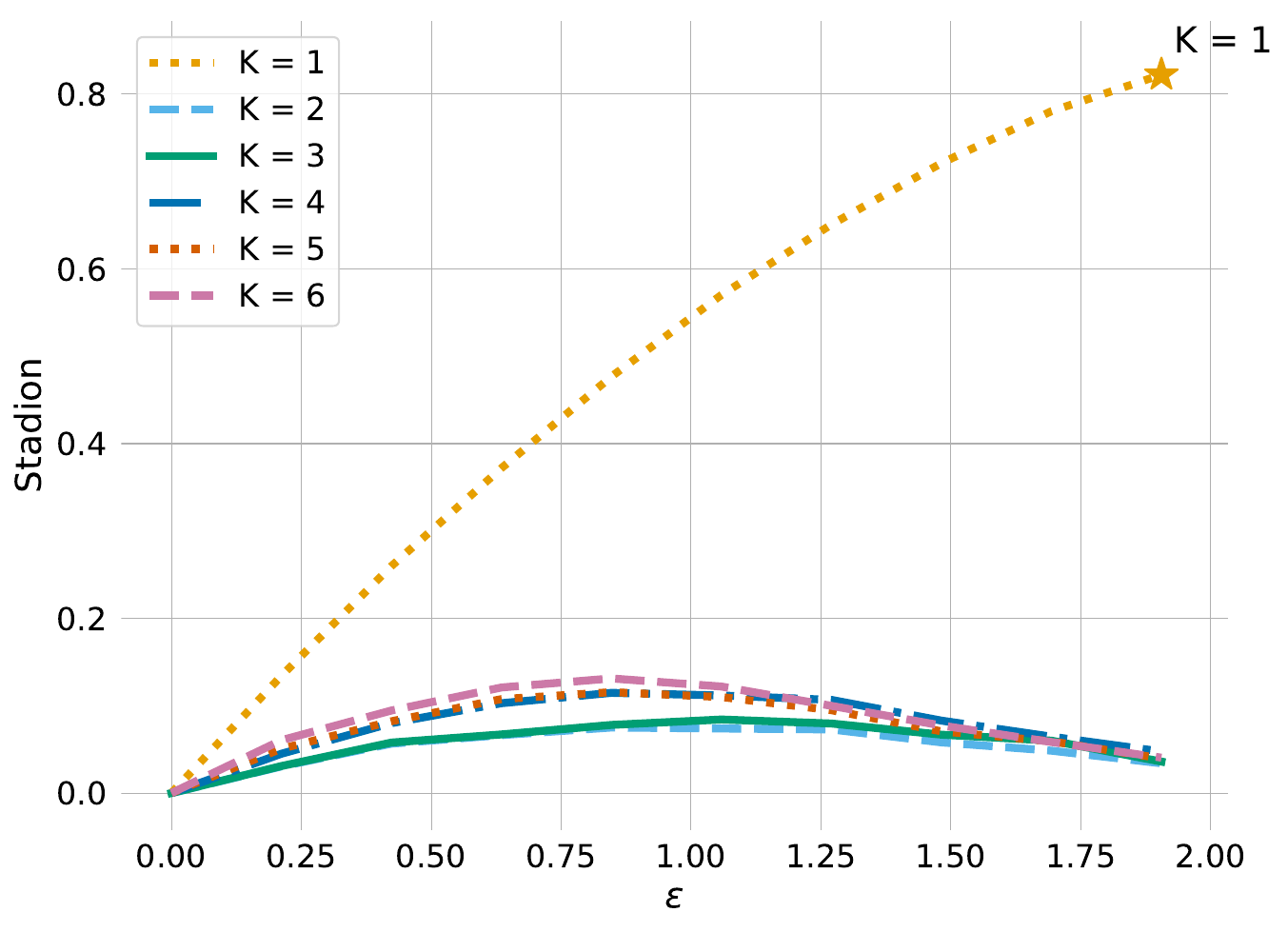}
     \includegraphics[width=0.49\columnwidth]{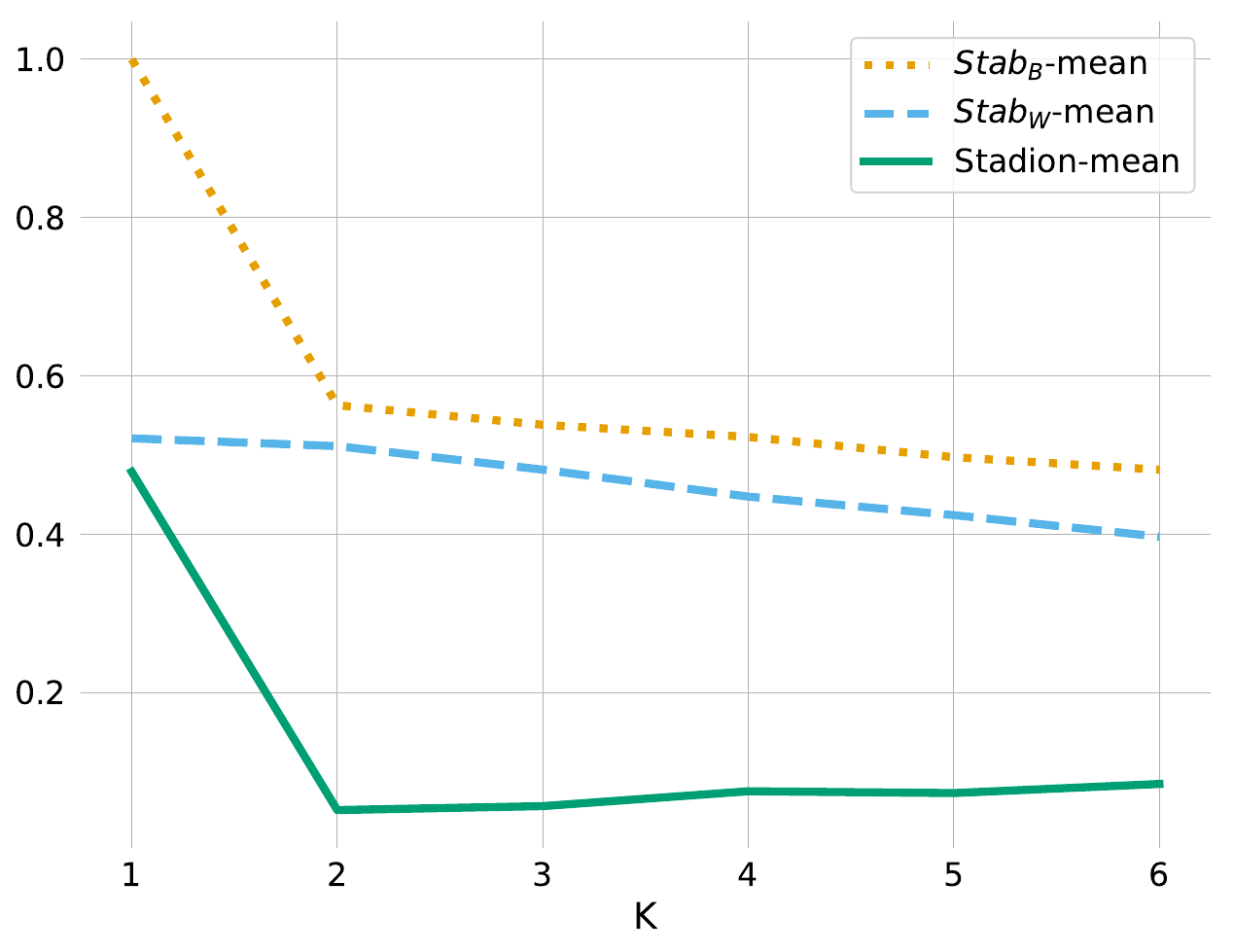}
    \caption{Stadion path (left) and stability trade-off plot (right) on the golfball data set with $K$-means. $K=1$ is clearly selected by Stadion-max/mean (uniform noise, $\mathbf{\Omega} = \{2, \ldots, 10\}$).}
    \label{fig:k1-golfball}
\end{figure}

Is a data set clusterable? Between-cluster stability is unable to answer this question, as the solution with a single cluster is trivially stable. Some stability methods are not even defined for $K=1$ because of normalization \cite{Lange2004}. Moreover, many internal indices use between-cluster distances and are not defined for a single cluster neither. We verified empirically that our criterion consistently outputs $K=1$ in the case when the algorithm does not find any cluster structure.
Table~\ref{tab:K1} contains results for non-clusterable distributions. Stadion outputs $K=1$ in all cases. An example of Stadion path and trade-off curve for the golfball data set is provided in Figure~\ref{fig:k1-golfball} (results are similar for other data sets).
\begin{table*}[h]
    \caption{Number of clusters found by Stadion on non-clusterable distributions.}
    \label{tab:K1}
    \centering
    \begin{small}
    \begin{tabular}{lcccc}
    \toprule
    Dataset & & N & dimension & $K$ selected by Stadion \\ 
    \midrule
    Uniform cube (2d) & \includegraphics[width=1.5cm]{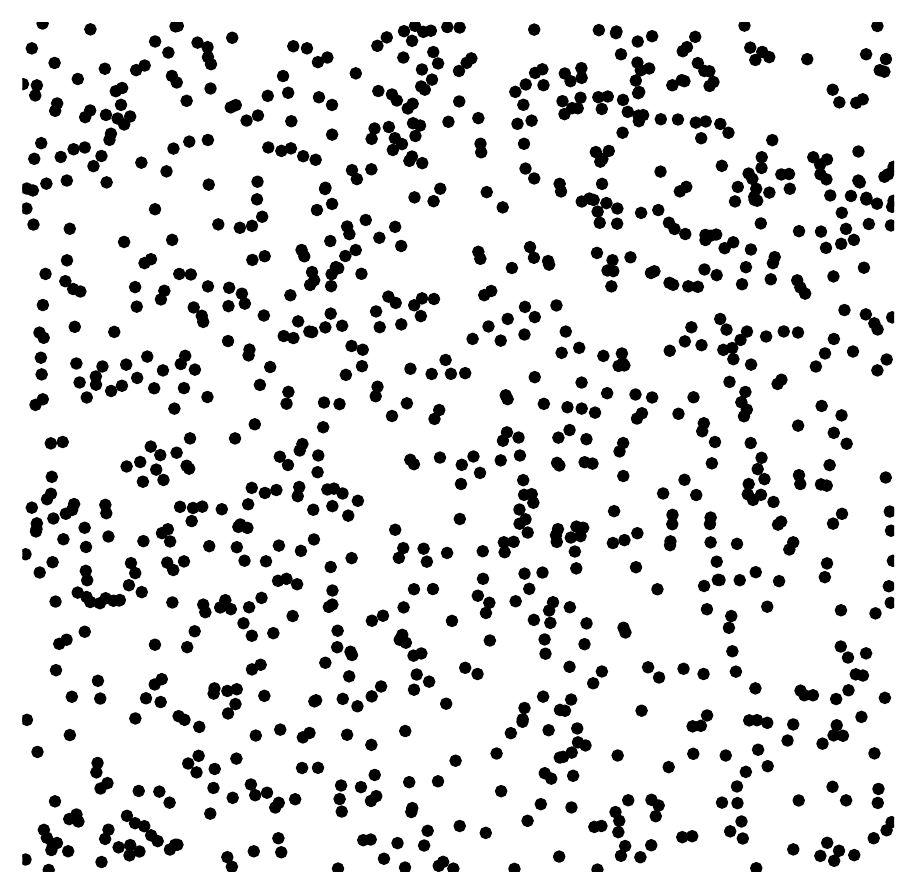} & 1000 & 2 & 1 \\
    Uniform cube (10d) &  & 1000 & 10 & 1 \\
    Gaussian (2d) & \includegraphics[width=2cm]{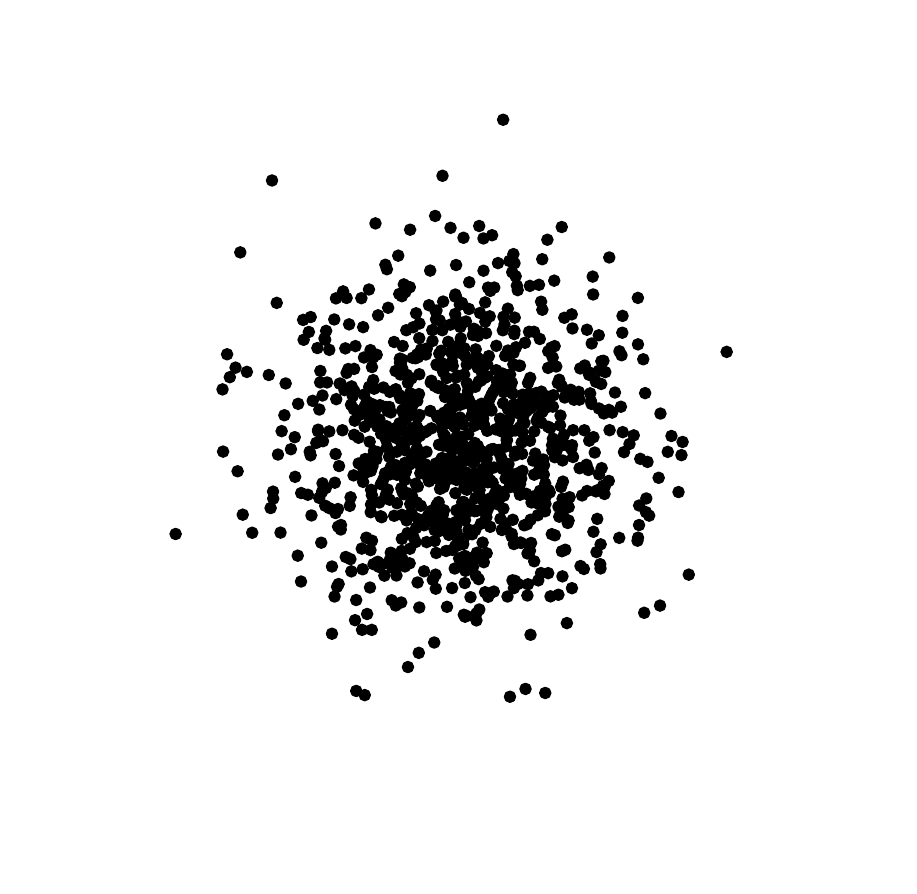} & 1000 & 2 & 1 \\
    Gaussian (10d) &  & 1000 & 10 & 1 \\
    Golfball \cite{ultsch2005} & \includegraphics[width=2cm]{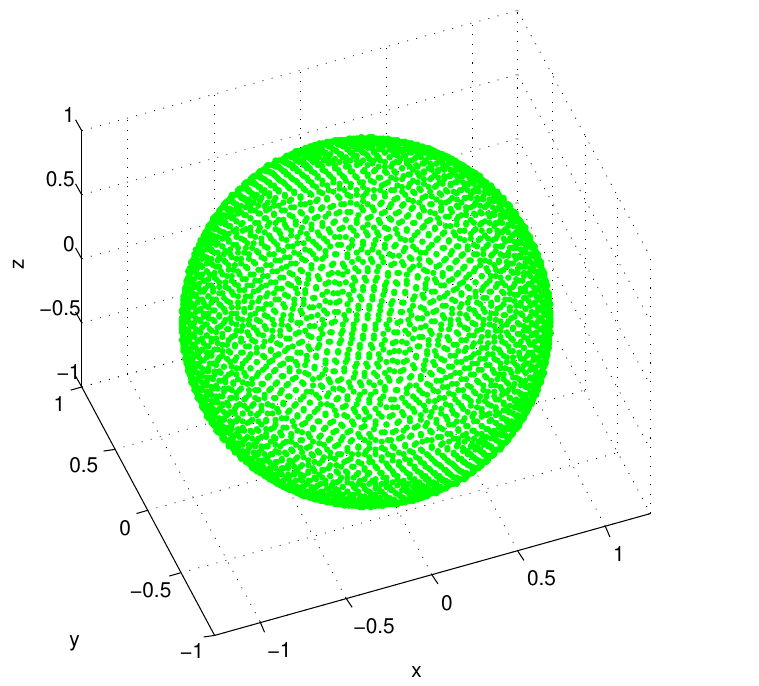} & 4002 & 3 & 1 \\
    \bottomrule
    \end{tabular}
    \end{small}
\end{table*}

\subsection{Examples of jumping between local minima}

\begin{figure*}[h]
    \centering
    \subfigure[]{\includegraphics[width=0.3\linewidth]{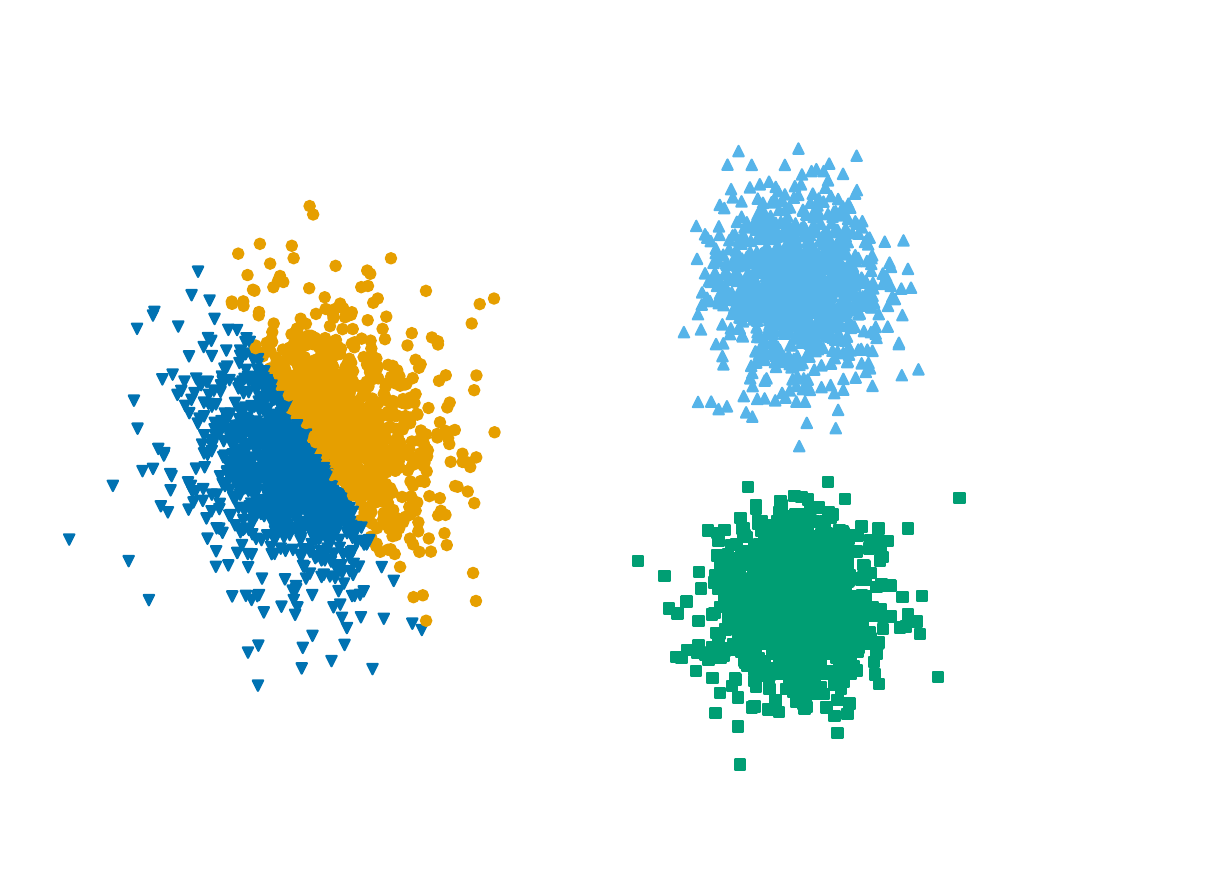}}
    \subfigure[]{\includegraphics[width=0.3\linewidth]{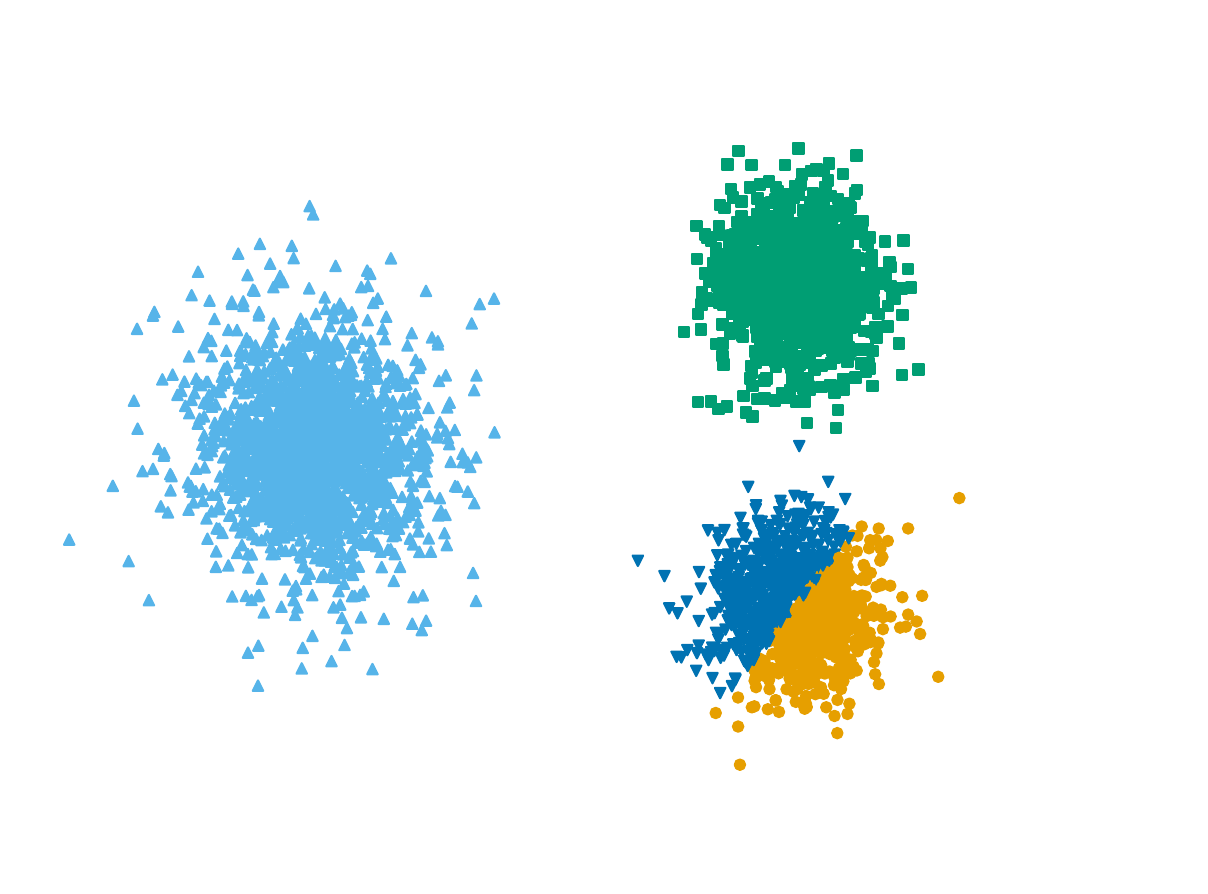}}
    \subfigure[]{\includegraphics[width=0.3\linewidth]{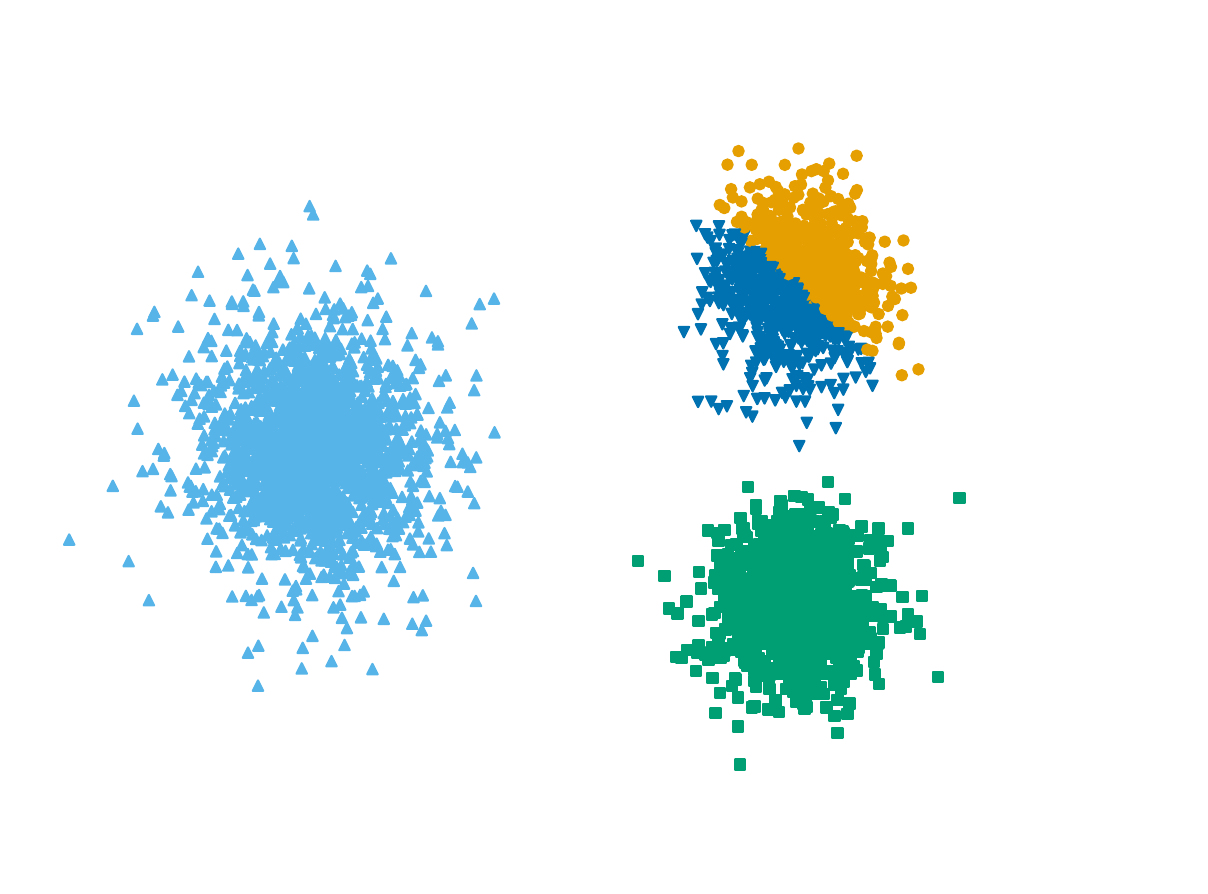}}
    \caption{Example of $K$-means jumping between three local minima for $K=4$, when a single random initialization is used. Depending on the initial centers configuration, the algorithm jumps between splitting a random cluster in two (a, b, c).} 
    \label{fig:jump-init}
\end{figure*}

As explained in Section 3, two sources of instability are jumping and jittering. We have already stated that our method leverages jittering of cluster boundaries in high-density regions due to perturbation. Jumping, on the other hand, happens when the algorithm finds very different solutions on different samples; in case of objective-minimizing algorithms, it ends up in different local minima. Two main effects lead to jumping: initialization, and symmetries in the data distribution. Finally, subtle geometrical properties of the distribution might also cause jumping \cite{von2010clustering}.
\begin{figure*}[h]
    \centering
    \includegraphics[width=0.3\linewidth]{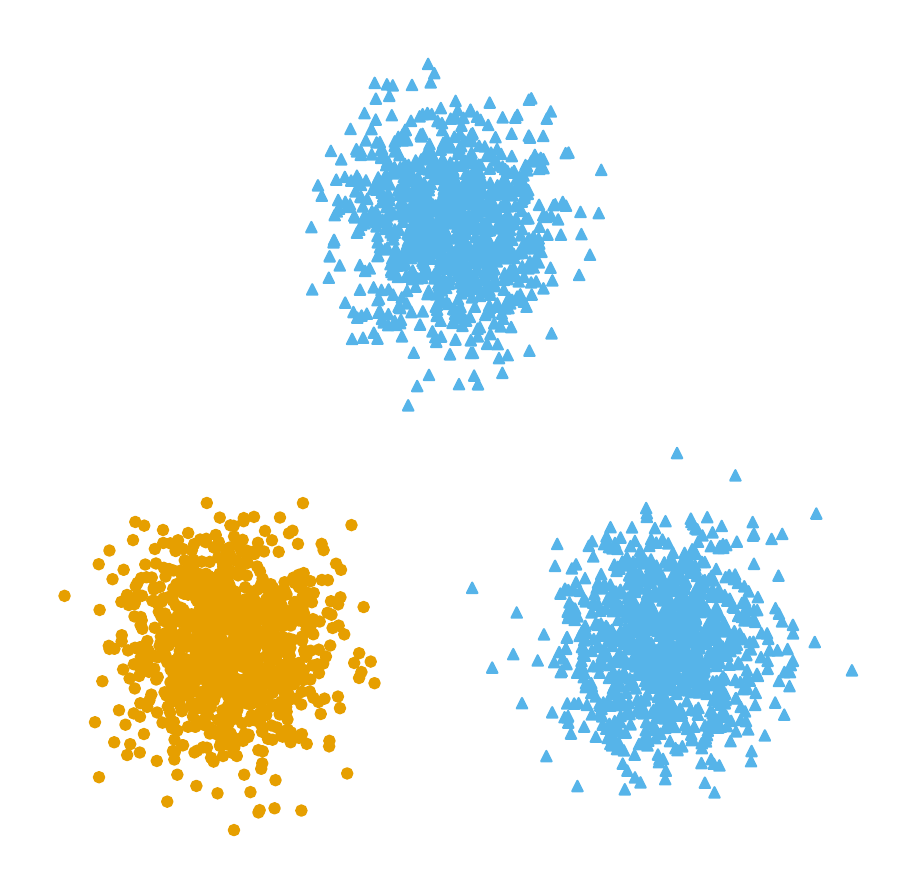}
    \includegraphics[width=0.3\linewidth]{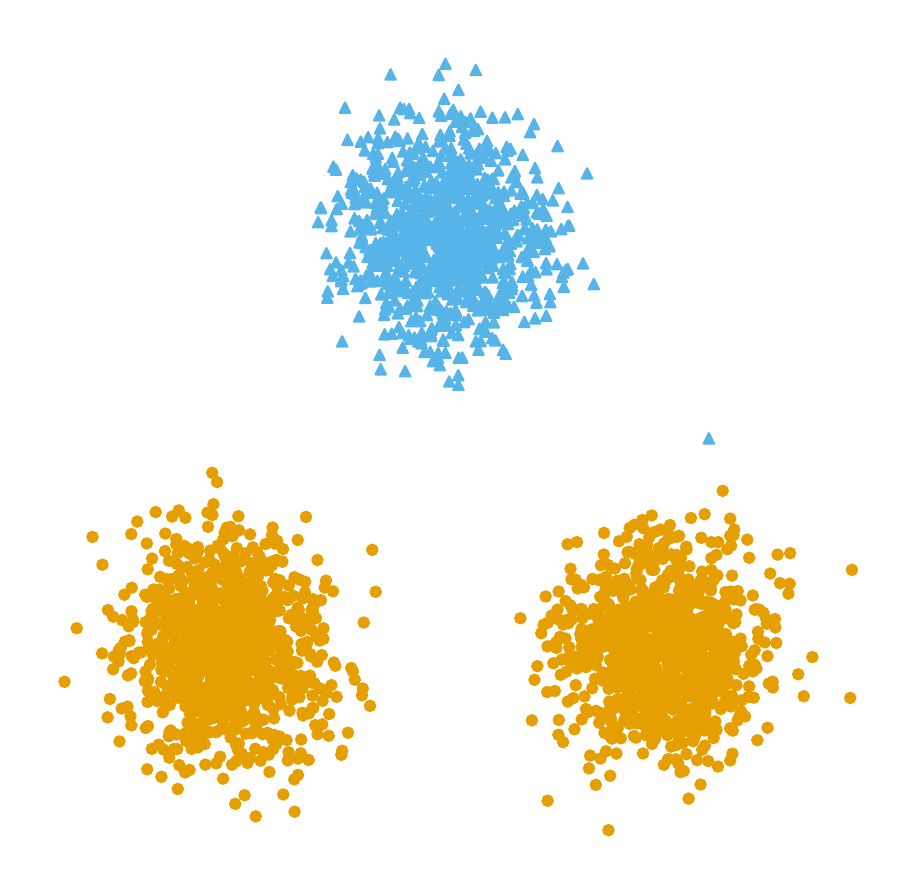}
    \includegraphics[width=0.3\linewidth]{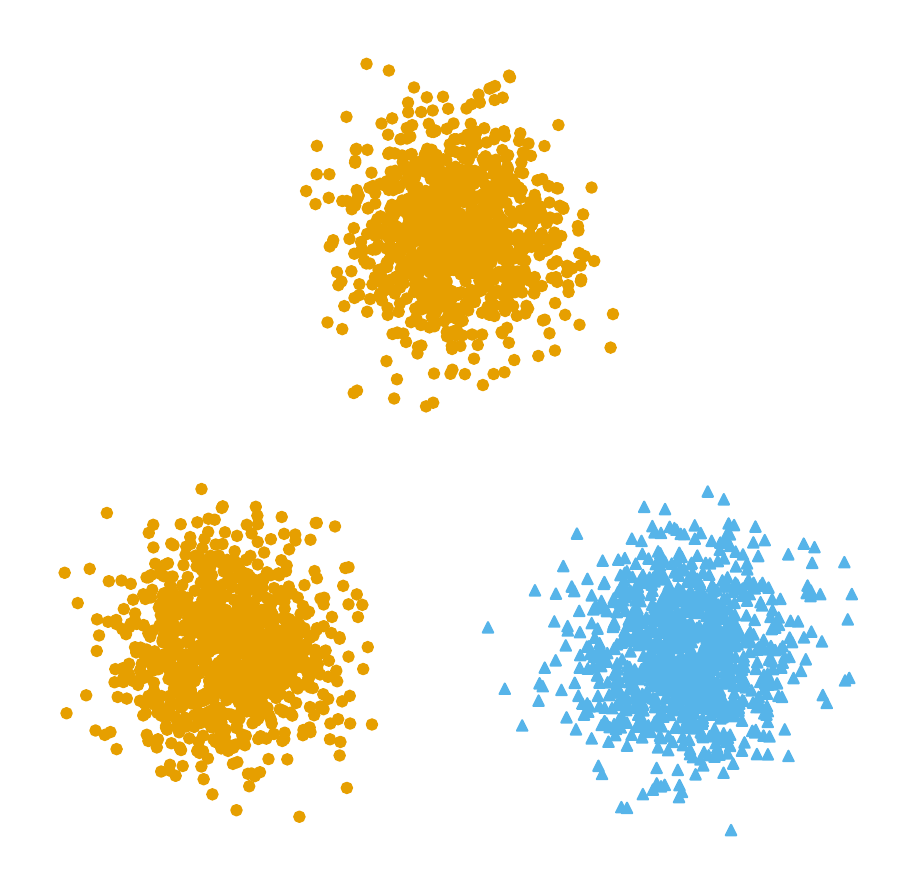}
    \caption{Example of $K$-means jumping between three global minima for $K=2$ on a symmetric distribution with three Gaussians, despite consistent initialization ($K$-means++ and best of 10 runs). Under slight perturbation (here uniform $\varepsilon$-AP, but resampling gives identical results), the algorithm jumps between grouping two random clusters together.}
    \label{fig:jump-sym}
\end{figure*}
An example of jumping of $K$-means due to symmetries is shown on Figure~\ref{fig:jump-sym}: clearly, there are several global minima, and even if the algorithm is deterministic, slight perturbations of the distribution (noise or sampling) make the algorithm jump between solutions. The second cause of jumping is due to initialization. As illustrated by Figure~\ref{fig:jump-init} for $K$-means, if a single random initialization is used, depending on the initial position of centers, four different configurations occur randomly, even without any perturbation of the data.
We place ourselves in a realistic setting without perfect symmetries and an effective algorithm initialization strategy, thus jumping is not the main source of instability.

\subsection{Failure of sampling-based stability methods}

\begin{figure*}[h]
    \centering
    \subfigure[]{\includegraphics[width=0.3\linewidth]{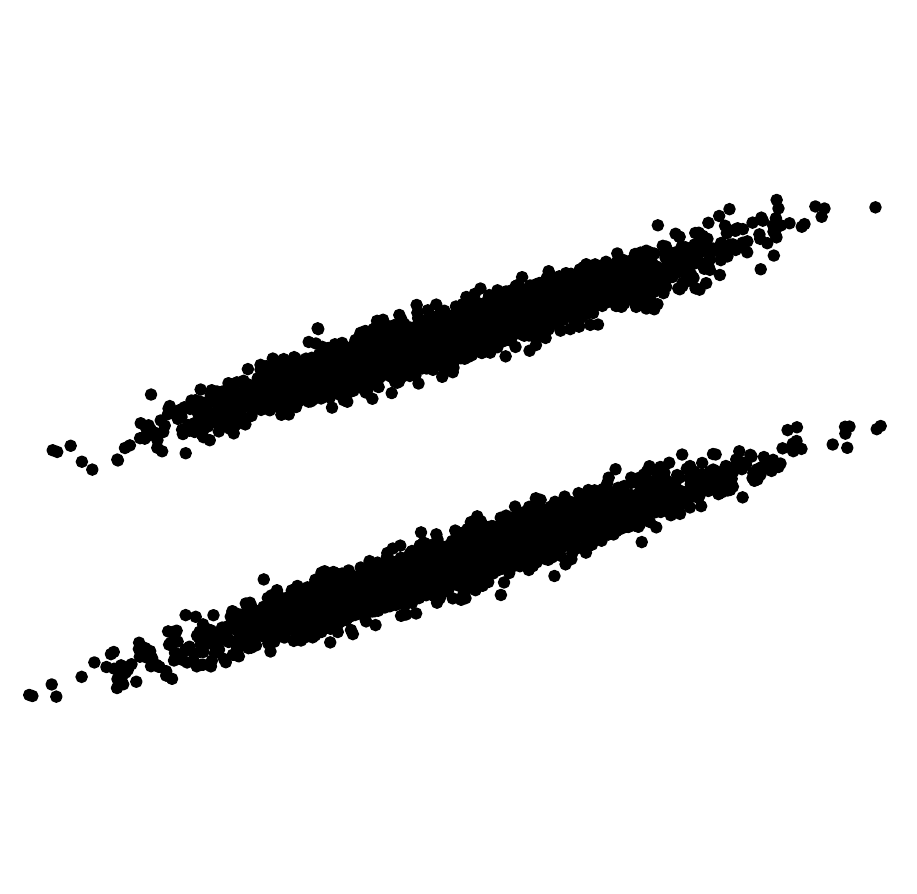}}
    \subfigure[$K=2$]{\includegraphics[width=0.3\linewidth]{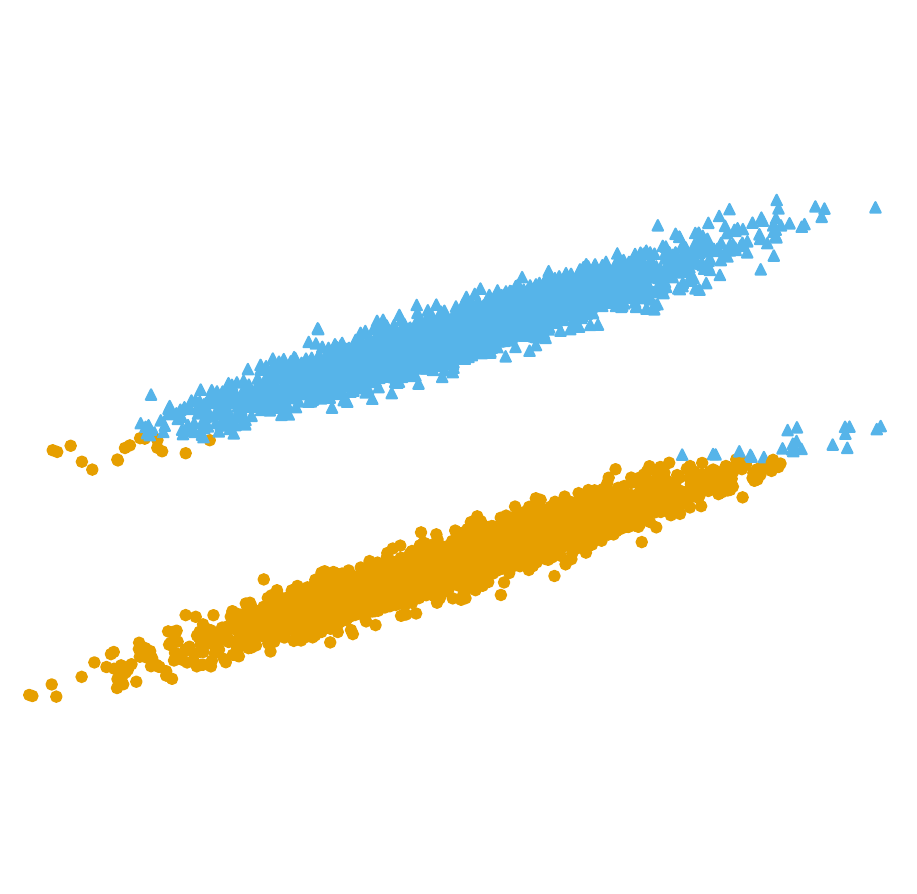}}
    \subfigure[$K=4$]{\includegraphics[width=0.3\linewidth]{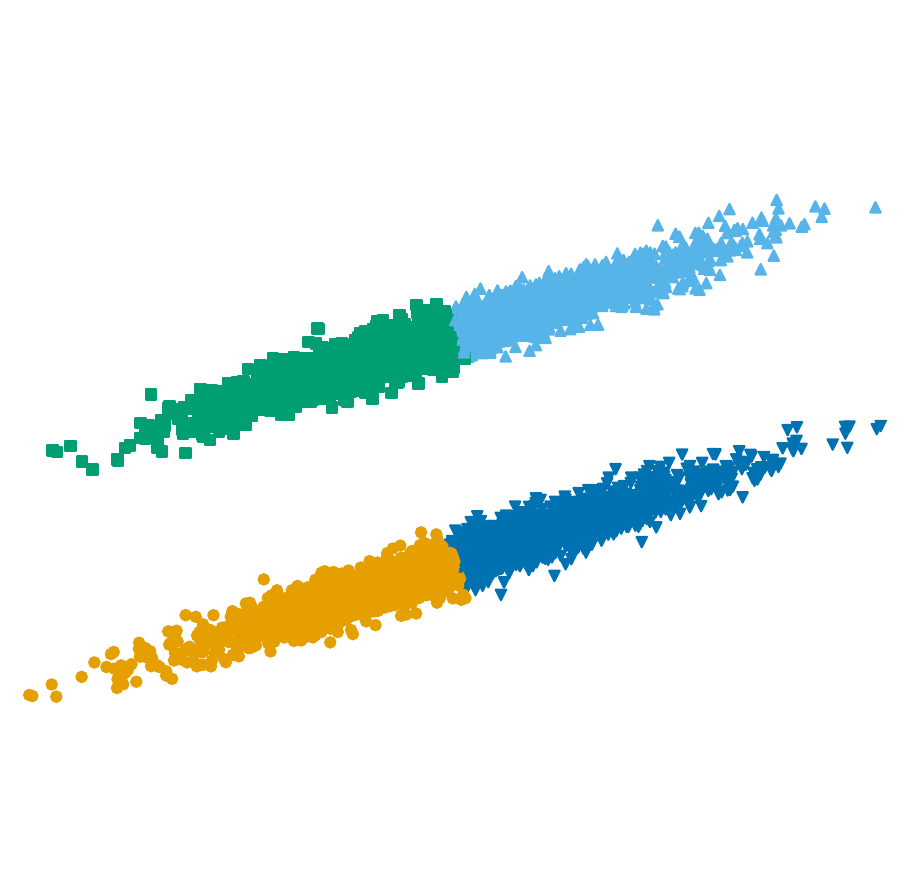}}
    \caption{Example data set of two correlated Gaussians, scaled to zero mean and unit variance. With the $K$-means algorithm, all sampling-based methods select $K=4$ or $K=6$, whereas with $\varepsilon$-Additive Perturbation, $K=2$ is the only stable solution.} 
    \label{fig:sampling-killer}
\end{figure*}

\begin{figure*}[h]
    \centering
    \includegraphics[width=0.32\linewidth]{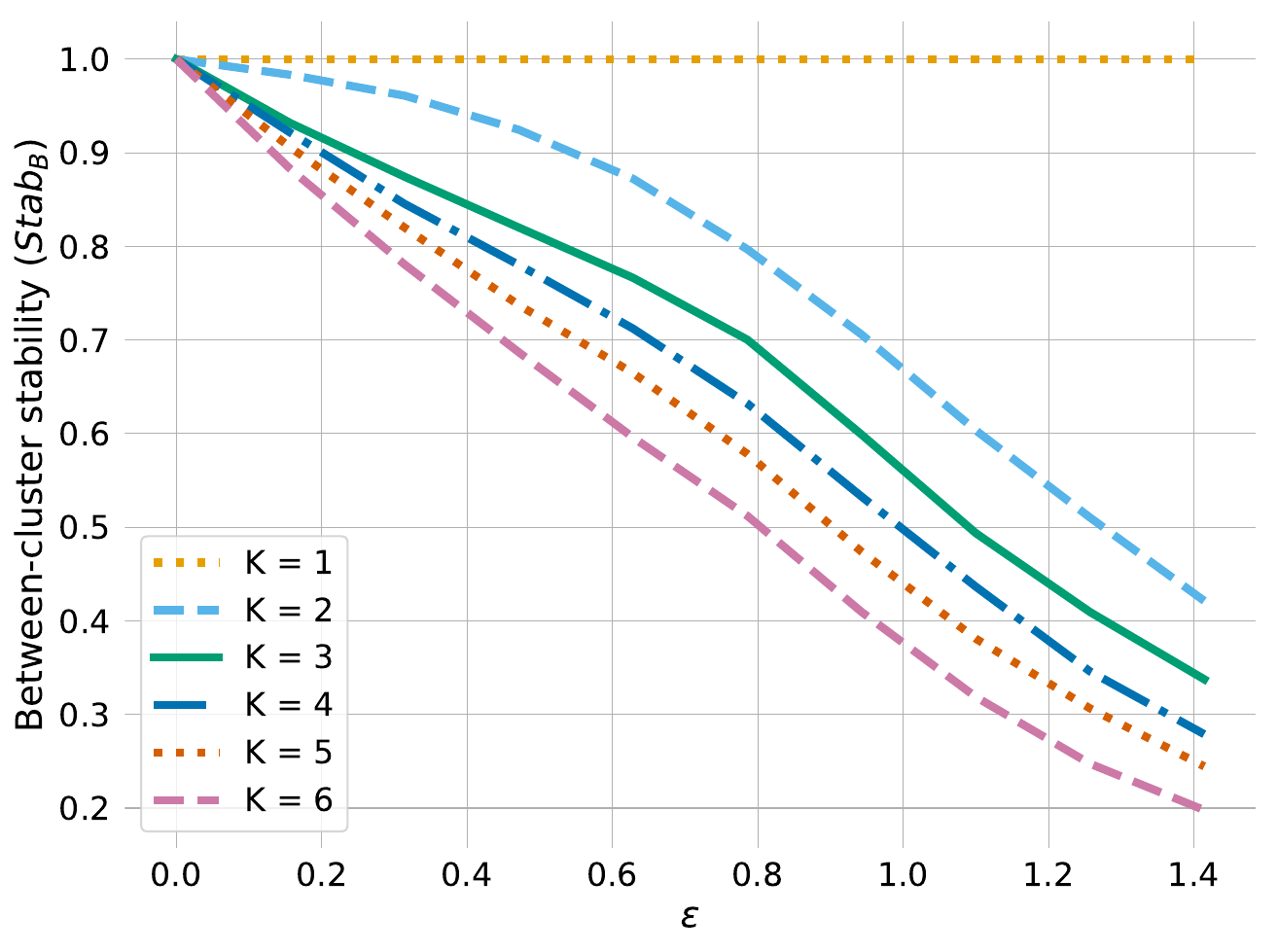}
    \includegraphics[width=0.32\linewidth]{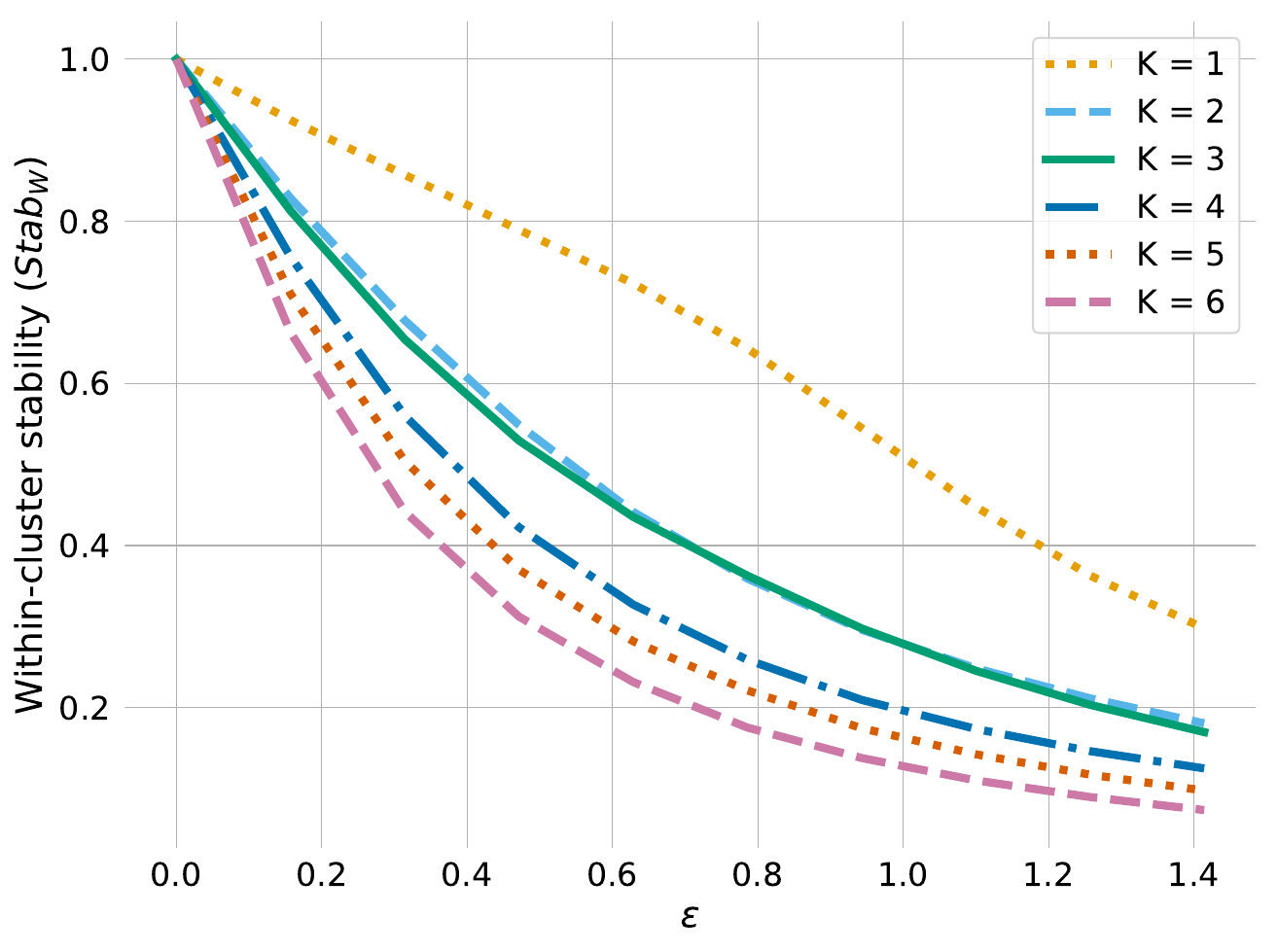}
    \includegraphics[width=0.32\linewidth]{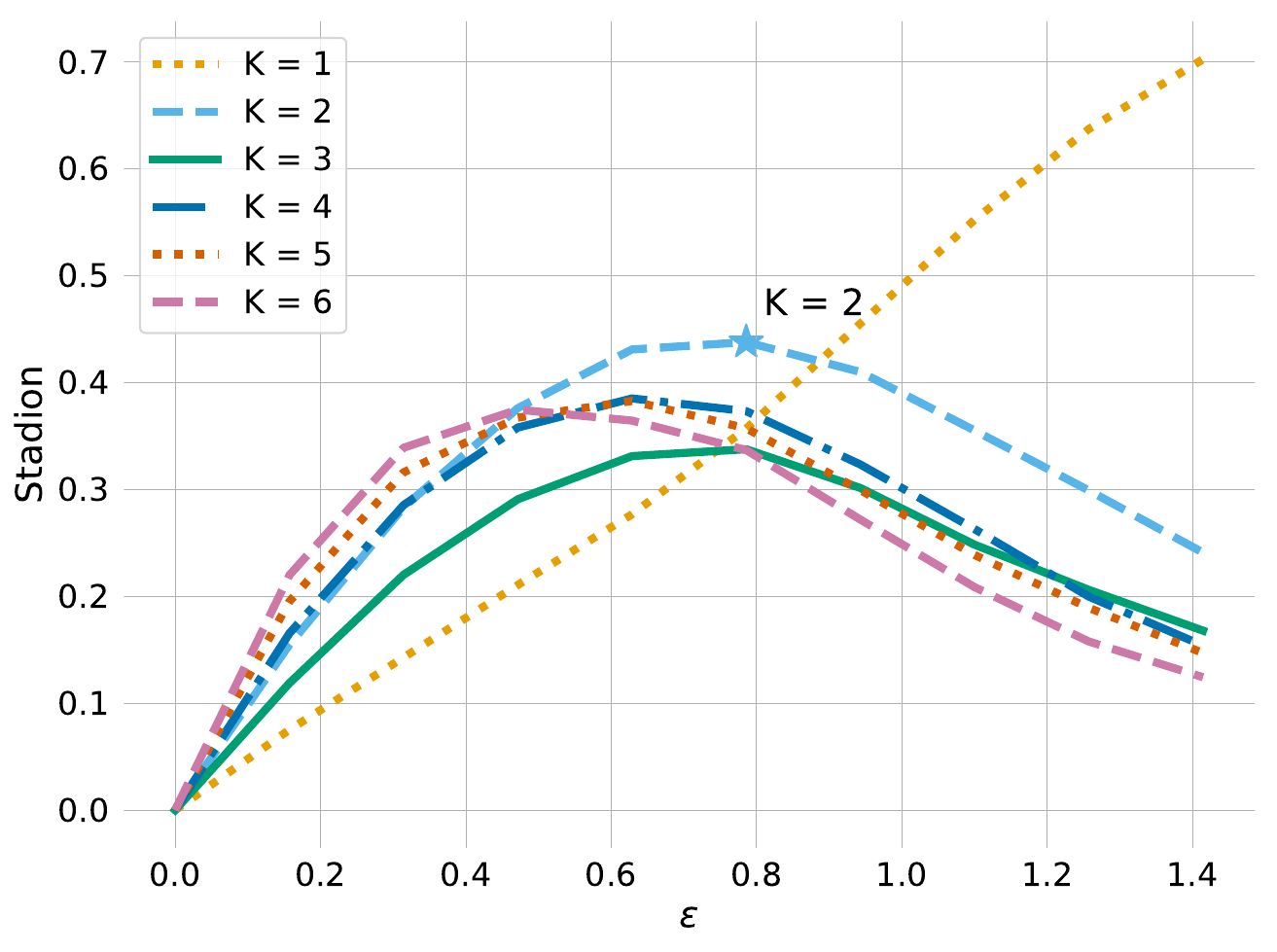}
    \caption{Between-cluster stability, within-cluster stability and Stadion paths (uniform noise, $\mathbf{\Omega} = \{2, \ldots, 6\}$) on the example of two correlated Gaussians where all sampling-based methods fail. Stadion clearly finds $K=2$ by taking the max or mean of the path curve.} 
    \label{fig:sampling-killer-paths}
\end{figure*}

In this section, we will see on a trivial example why stability methods based on re-sampling are not reliable to detect the presence of structure in the data. Four methods are compared:
\begin{enumerate}
    \item Stadion based on $\varepsilon$-Additive Perturbation
    \item Stadion based on bootstrapping perturbation
    \item The model explorer algorithm \cite{Ben-Hur2002} based on subsampling
    \item The model order selection method \cite{Lange2004} based on splitting data in two halves and transferring labels from one half onto the other using a supervised nearest-neighbor classifier.
\end{enumerate}
We demonstrate that only the first method is successful on a simple example consisting in a mixture of two correlated Gaussians, represented on Figure~\ref{fig:sampling-killer}. Data are scaled to zero mean and unit variance as for every other data set. $K$-means is used to cluster the data. As illustrated on the plot, $K$-means with $K=2$ separates almost perfectly the two Gaussians. All other solutions split the two Gaussians into several sub-clusters of equal sizes, with cluster boundaries lying in the regions of highest density, as can be seen from the example for $K=4$ (where the boundaries are in the middle of the Gaussians). Thus, in addition to being the best solution, $K=2$ is the only acceptable one. However, sampling-based methods fail in assessing its stability, since they estimate $K=4$ as the most stable solution. This result can be explained because the data set is not symmetric and for each $K$ there is one global minimum so no jumping occurs, even with a random initialization scheme. Thus the only possible source of instability stems from jittering. As expected in theory, our experiments show that the different sampling processes did not succeed in creating jittering. Conversely, $\varepsilon$-AP has indeed produced jittering.
In details, the model order selection method \cite{Lange2004} selects $K=4$, followed by $K=6$. The model explorer \cite{Ben-Hur2002} finds $K=6$ as the best solution, followed by $K=4$. These results are consistent across initialization schemes (random, $K$-means++, best of several runs). Hence, random initialization will not help creating instability by jumping. Stadion with additive noise was able to find $K=2$ among the set of tested values $\{1, \ldots, 6\}$ (see Figure~\ref{fig:sampling-killer-paths}). This is not only due to adding the within-cluster stability. As evidence, we replaced $\varepsilon$-AP by a bootstrap perturbation: Stadion with bootstrapping also fails, selecting $K=1$ as the best solution followed by $K=4$, and this for all initialization schemes.

\subsection{Example of Stadion behavior with $K$-means}

\begin{figure}[h]
    \centering
    \includegraphics[width=0.4\linewidth]{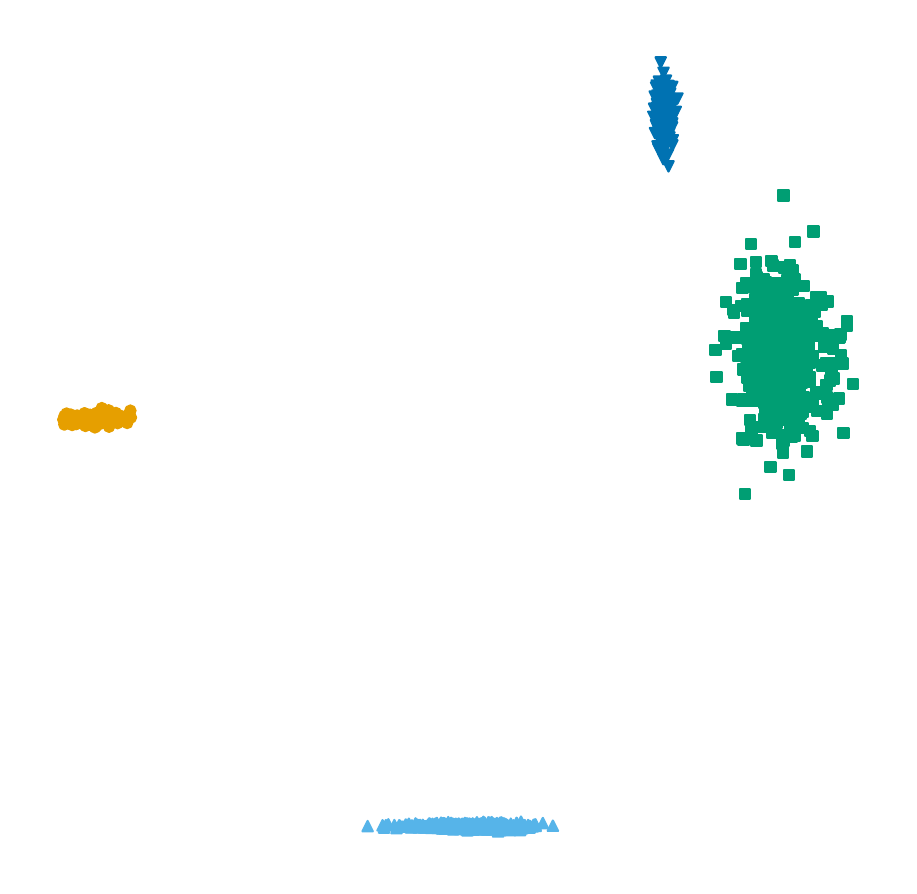}\hfill
    \includegraphics[width=0.49\linewidth]{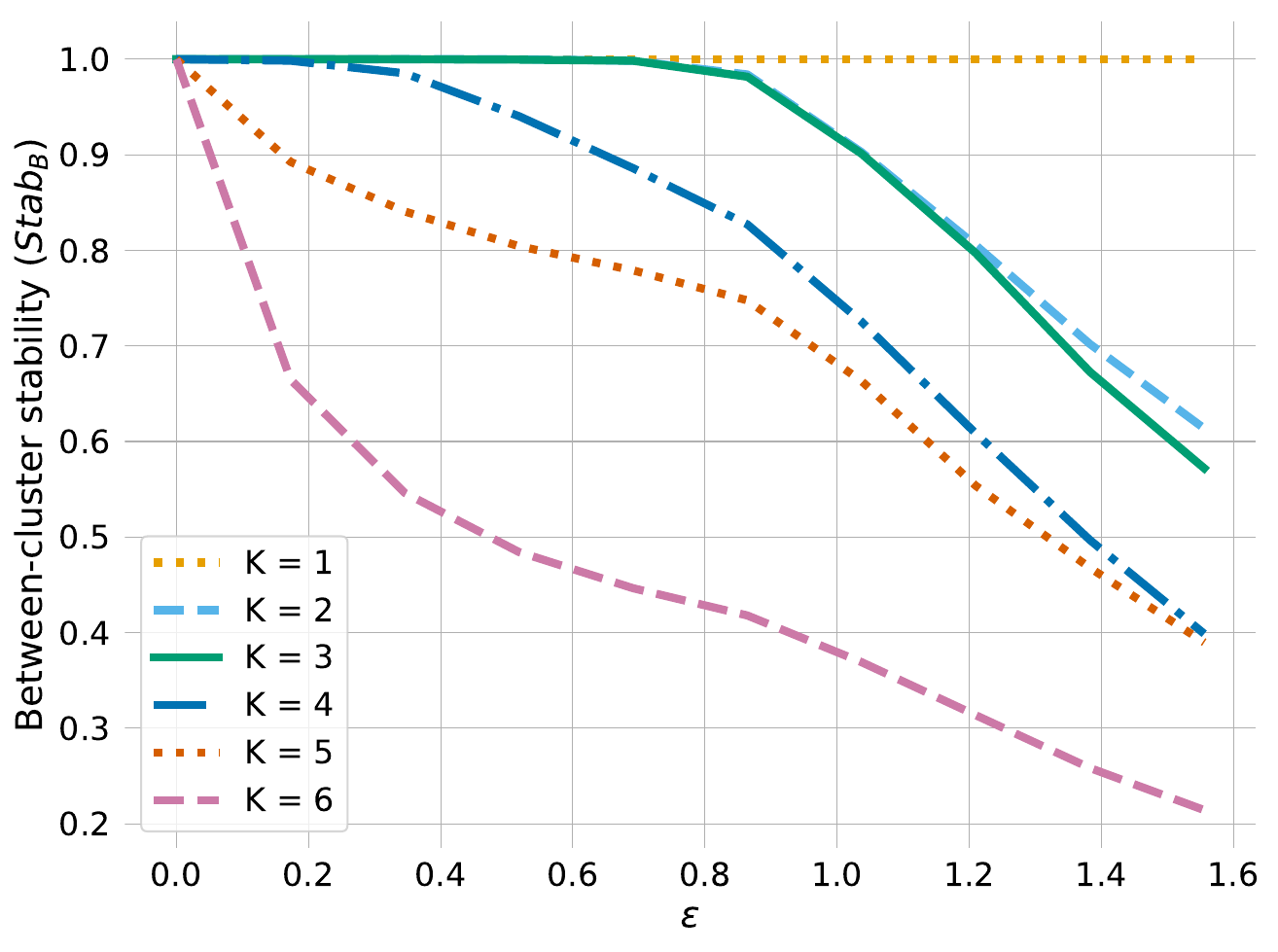}\\
    \includegraphics[width=0.49\linewidth]{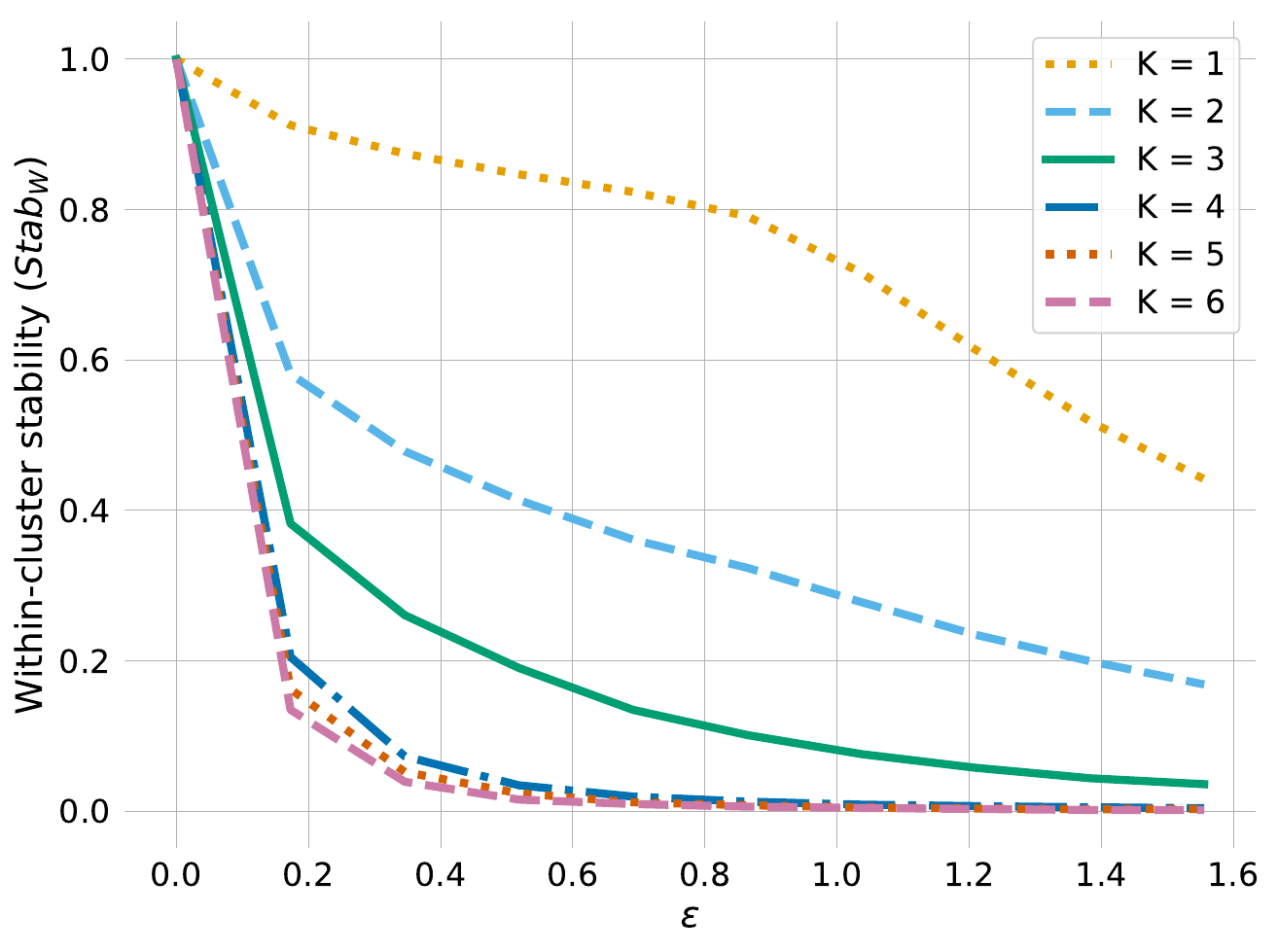}
    \includegraphics[width=0.49\linewidth]{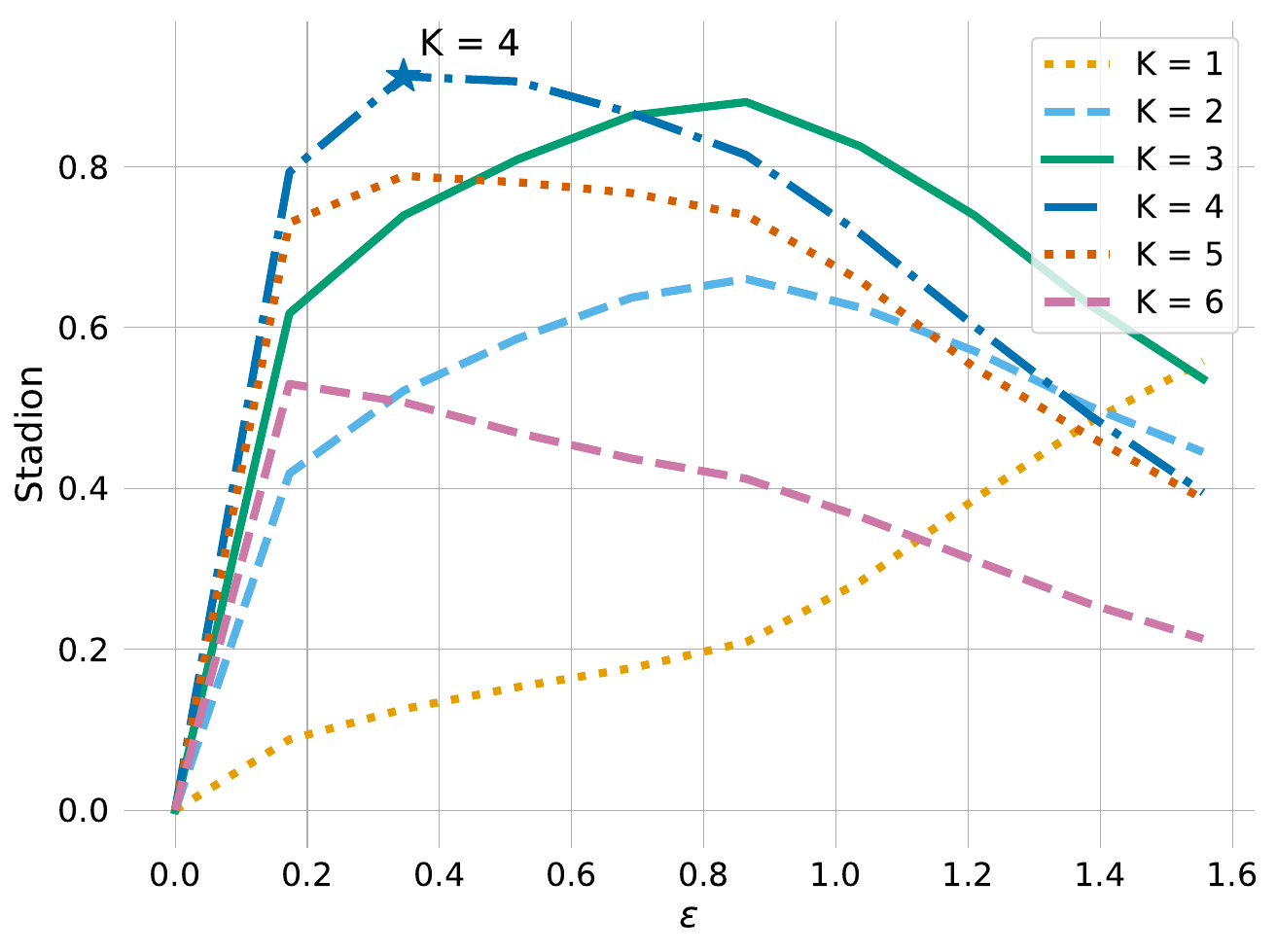}
    \caption{Between-cluster stability, within-cluster stability and Stadion paths (uniform noise, $\mathbf{\Omega} = \{2, \ldots, 6\}$) on the 2d-4c data set. Stadion selects $K=4$, followed by $K=3$.}
    \label{fig:demo-ex1}
\end{figure}

This example illustrates the behavior of our stability criterion Stadion and how to interpret the stability paths, using the data set 2d-4c shown in Figure~\ref{fig:demo-ex1}. It consists in four clusters with different variance and size, where two clusters are closer to each other while the other clusters are at a greater distance. At first glance, this example looks trivial, but the majority of internal indices fail. For instance, the Dunn and Silhouette indices both select $K=3$.
The stability paths in Figure~\ref{fig:demo-ex1} show that Stadion is able to detect the structure of the data and selects $K=4$. The only difference between the solutions with $K=4$ and $K=5$ is that the largest cluster (in green) is split, thus leading to a much lower between-cluster stability but the same within-cluster stability. Solutions $K=2$ and $K=3$ group clusters together without any splitting. Therefore, those solutions have a high between-cluster stability and also a high within-cluster stability. Altogether, on the Stadion path (Figure~\ref{fig:demo-expath1}), the path corresponding to $K=4$ is similar to $K=5$ whereas $K=2$ and $K=3$ have an equivalent behavior. This is due to the structure of the data, and especially because the two rightmost clusters are close to each other. The moment when the path of solution $K=3$ becomes the best solution is the moment when these two clusters merge because of a high $\varepsilon$-AP, and this is also the moment where $K=1$ prevails.

Finally, Stadion paths (with stability and instability paths) give useful additional information on a clustering and on the structure of the data. When $K > K^{\star}$, the paths are similar to the path of $K^{\star}$ but with a smaller scale, as they have the same within-cluster stability but lower between-cluster stability. On the other hand, when $K < K^{\star}$, the paths are shifted towards the right, and may become superior for larger $\varepsilon$ values.

\subsection{Whenever $K^{\star}$ is not the best partition}

\begin{figure*}[h]
    \centering
    \subfigure[$K=2$]{\includegraphics[width=0.24\linewidth]{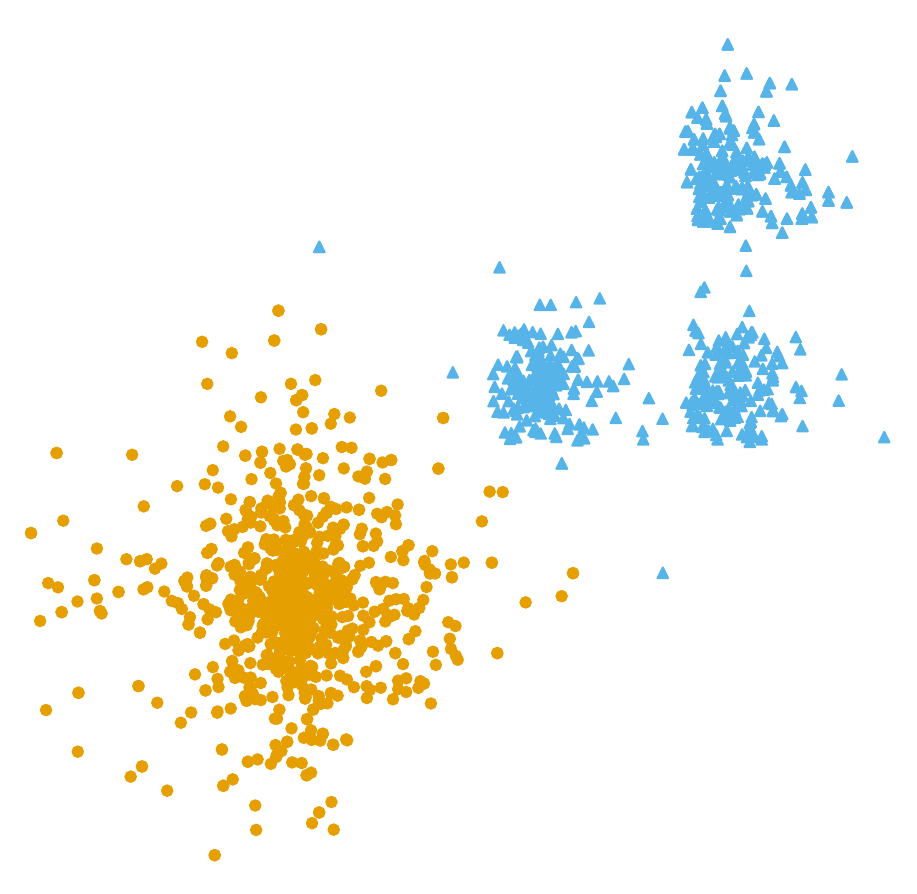}}
    \subfigure[$K = 3$]{\includegraphics[width=0.24\linewidth]{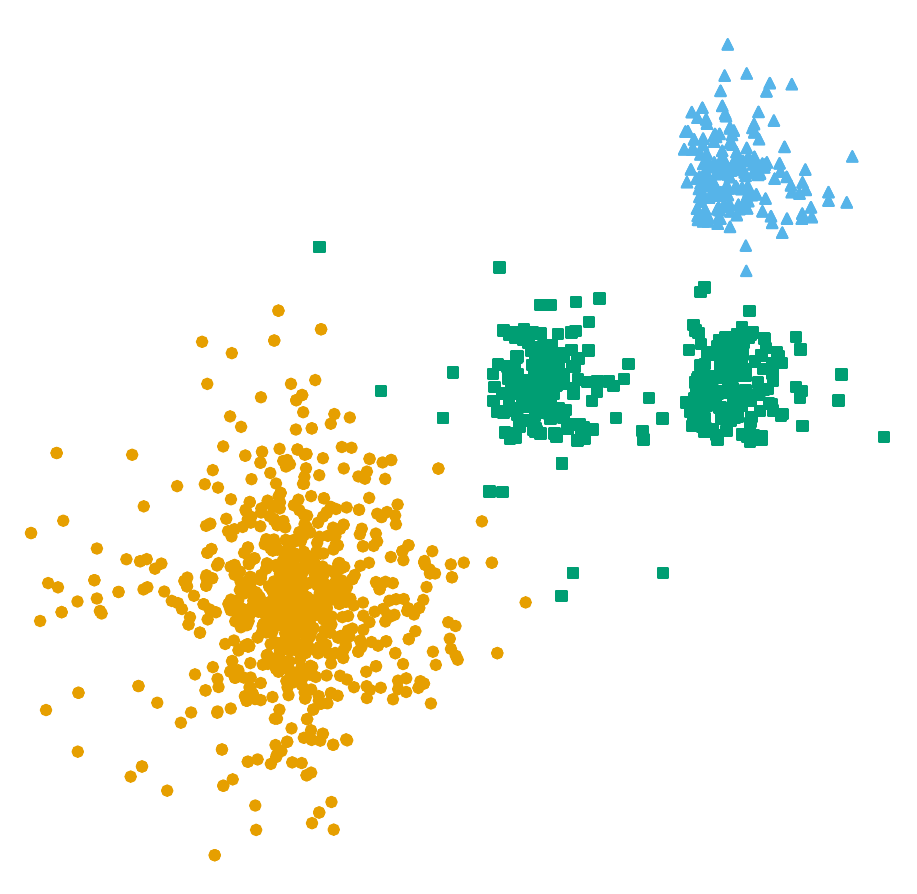}}
    \subfigure[$K^{\star} = 4$]{\includegraphics[width=0.24\linewidth]{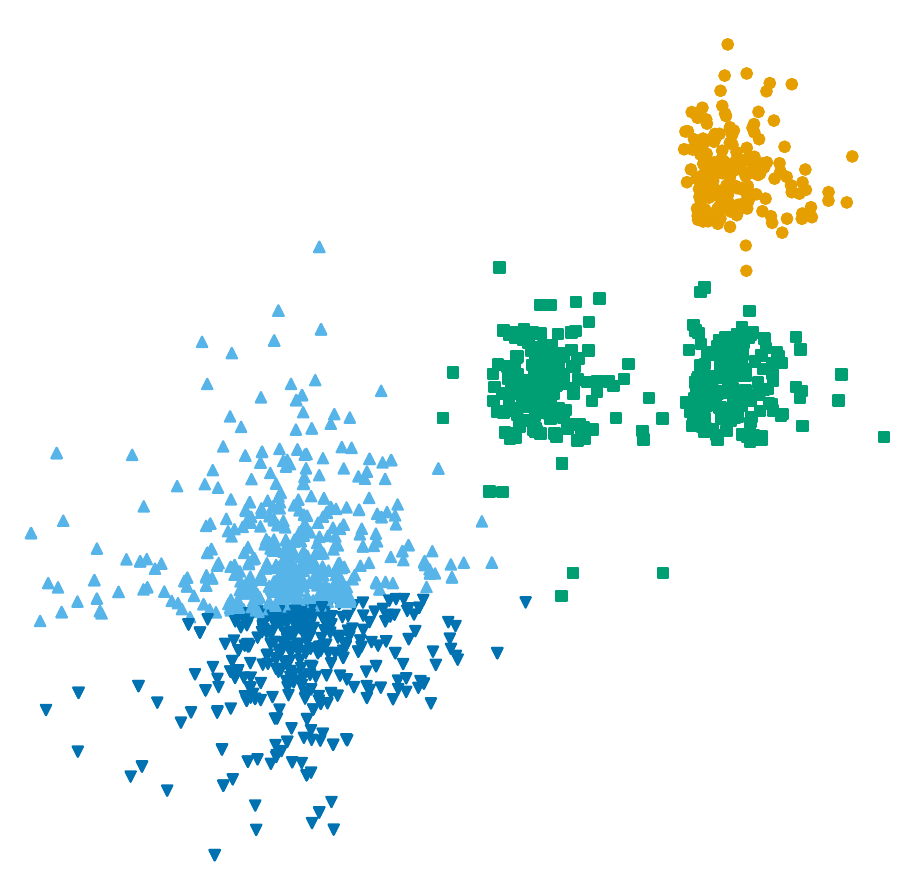}}
    \subfigure[$K = 5$]{\includegraphics[width=0.24\linewidth]{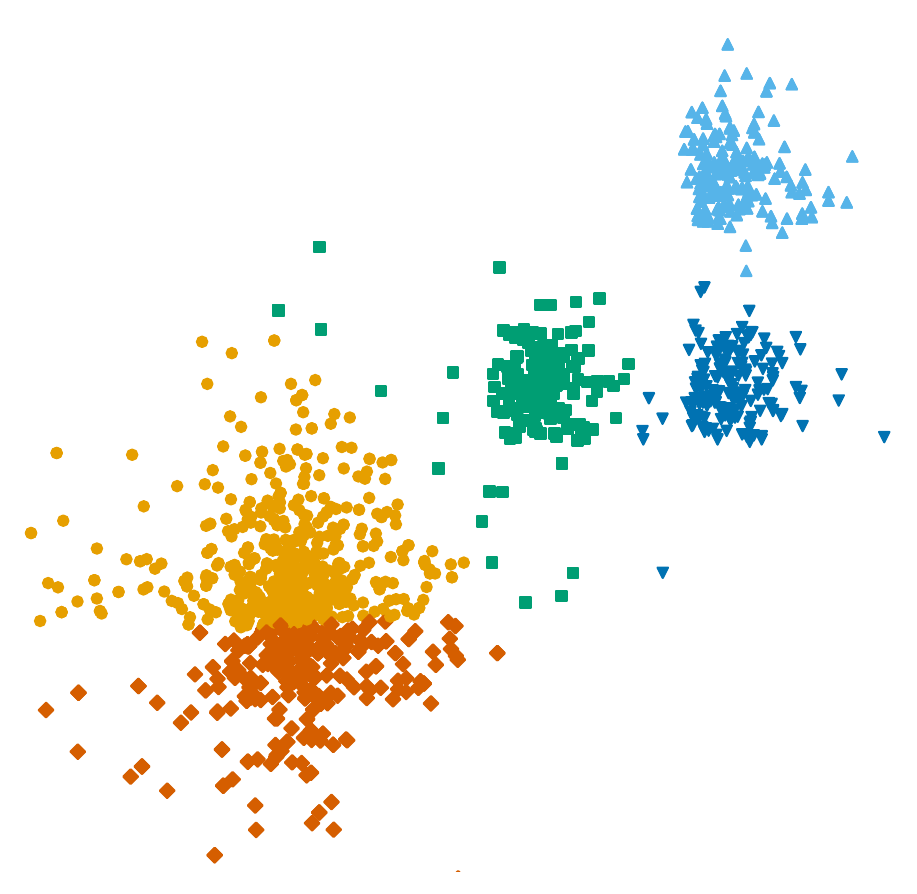}}
    \caption{Solutions of $K$-means on the 4clusters\_corner data set for $K \in \{2, \ldots, 5\}$.}
    \label{fig:4clusters-corner}
\end{figure*}

\begin{figure}
    \centering
    \includegraphics[width=0.6\linewidth]{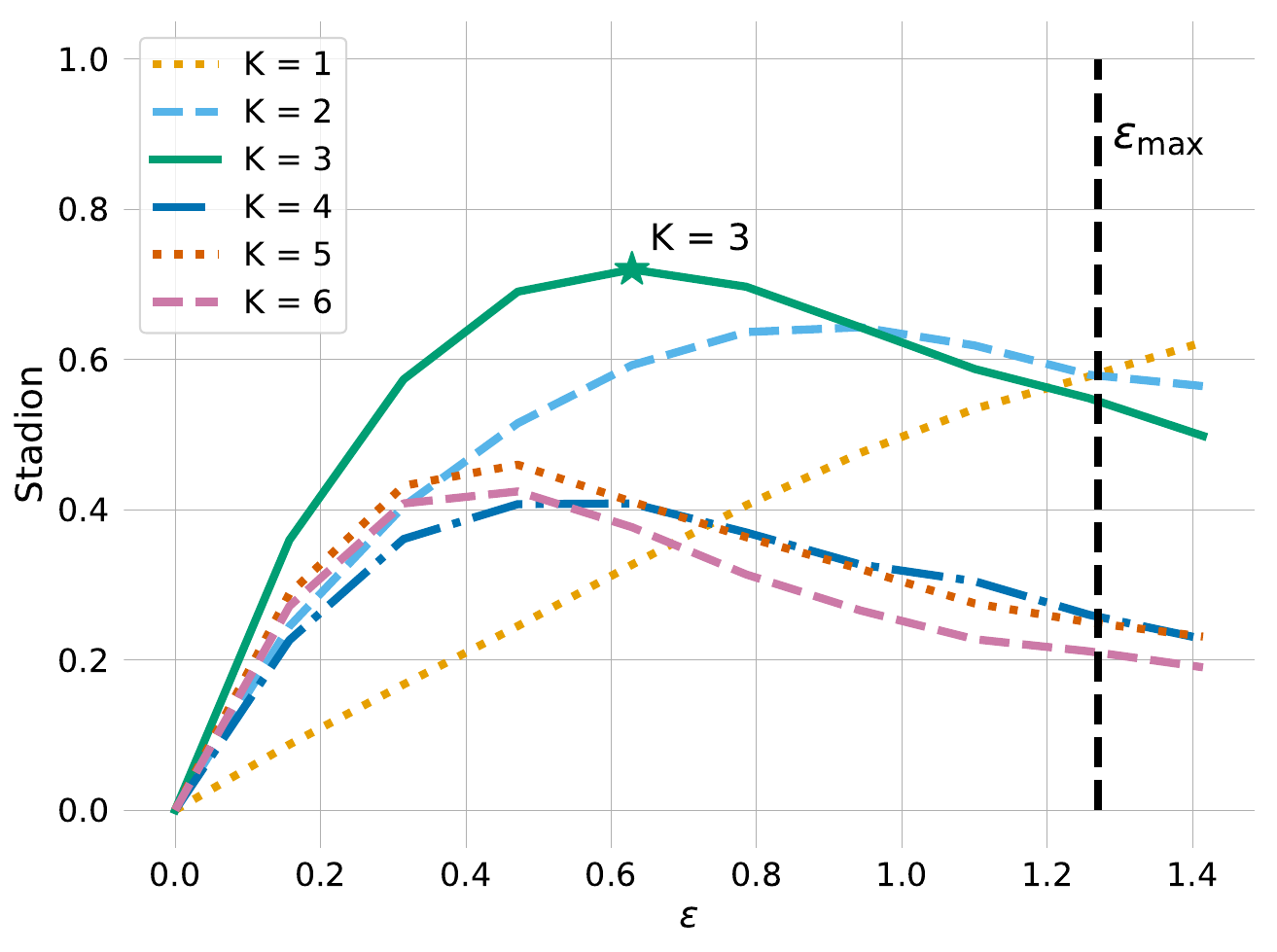}
    \caption{Stadion paths on the 4clusters\_corner data. $K=3$ is selected although $K^{\star} = 4$.}
    \label{fig:4clusters_corner-paths}
\end{figure}

Sometimes, the best solution is not the one obtained with the true parameter $K^{\star}$, because the algorithm itself is unable to recover the ground-truth partition. This is the case for the 4clusters\_corner data set, depicted on Figure~\ref{fig:4clusters-corner}. While the most natural solution is to separate the four clusters, it is not achievable by $K$-means: with $K^{\star} = 4$, it will cut through the large cluster instead of separating the two small green clusters, for the sake of saving the cost induced by the variance and the size of this cluster. Among the proposed solutions, the highest ARI (w.r.t. the ground-truth) is obtained with $K = 3$ (ARI = 0.92), followed by $K = 2$ (0.74), $K = 5$ (0.65) and lastly $K^{\star} = 4$ (0.58).
Almost all internal indices select $K = 2$. Stability methods based on sampling \cite{Ben-Hur2002,Lange2004} selected the ground-truth $K^{\star} = 4$, earning them a "win", although it is the worst partition among the four. We explain it by the fact that these methods are not leveraging jittering inside the large cluster. Finally, Stadion always selects the solution $K = 3$ having the highest ARI. Moreover, the criterion outputs solutions in the same order than ARI. This examples clearly exhibits the stability trade-off occurring in Stadion: it tries to preserve a high between-cluster stability while keeping within-cluster stability as low as possible (see Table~\ref{tab:4clusters_corner}). Stadion paths on Figure~\ref{fig:4clusters_corner-paths} also show how the three smaller clusters merge as the noise level increases.

\begin{table}[h]
    \centering
    \caption{Stability trade-off leveraged by Stadion on the 4clusters\_corner data set.}
    \label{tab:4clusters_corner}
    \begin{tabular}{ccccc}
    \toprule
    K & ARI & StabB & StabW & Stadion \\
    \midrule
    1 & 0.00 & ++  & - - & \; 0 \; \XSolidBrush \\
    2 & 0.74 & ++  & -   & \; + \; \XSolidBrush \\
    3 & 0.92 & ++  & +   & +++ \Checkmark \\
    4 & 0.58 & - - & +   & \; - \; \XSolidBrush \\
    5 & 0.65 & - - & ++  & \; 0 \; \XSolidBrush \\
    \bottomrule
    \end{tabular}
\end{table}
%
\subsection{Example of $K$ approaching $N$}

\begin{figure}[h]
    \centering
    \includegraphics[width=0.6\linewidth]{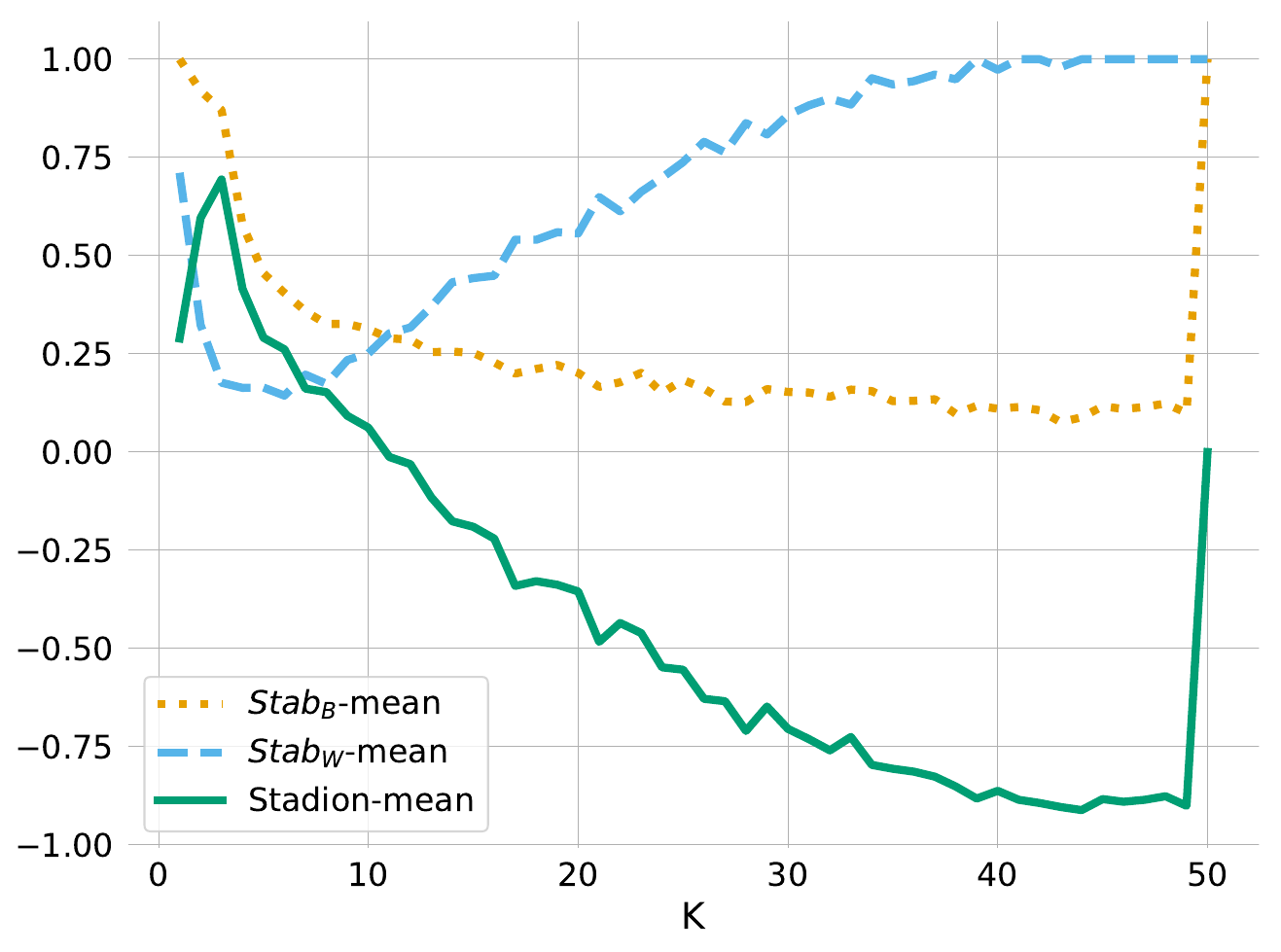}
    \caption{Stability trade-off plot for $K$-means on three Gaussians with $N = 50$ for $K \in \{1, \ldots, 50\}$ (uniform noise, $\mathbf{\Omega} = \{2, \ldots, 10\}$). Stadion is still valid when the tested $K$ becomes large.}
    \label{fig:trade-off-N}
\end{figure}

This paragraph introduces the behavior of Stadion when the number of clusters $K$ evaluated becomes as large as the number of samples $N$. Even if this is beyond the common setting in clustering, the criterion is still consistent. Figure~\ref{fig:trade-off-N} displays the stability trade-off for $K$-means on an example with three Gaussians, using ARI as the similarity metric fot stability estimation. As $K$ approaches $N$,
\begin{enumerate}
    \item Between-cluster stability decreases towards $0$, except for $K=N$ where it jumps back to $1$, because all partitions with one sample per cluster are perfectly similar to ARI.
    \item Within-cluster stability increases towards $1$, as clusters with few samples become trivially stable.
    \item Stadion still indicates the correct solution $K=3$, while decreasing towards $-1$, only jumping back to $0$ when $K=N$.
\end{enumerate}
Note that the borderline case $K = N$ depends on the similarity measure $s$ used.

%% file: supplementary/influence.tex
\section{Hyperparameter study}\label{appendix:influence}

The stability difference criterion (Stadion) introduced in this work is governed by several hyperparameters:
\begin{itemize}
    \item $D$: the number of perturbed samples used in the stability computation.
    \item noise: the type of noise for $\varepsilon$-Additive Perturbation. We experimented with uniform and Gaussian noise.
    \item $\mathbf{\Omega}$: the set of algorithm parameters $K'$ used in within-cluster stability.
\end{itemize}
The goal of this section is to study their importance and impact on the performance of Stadion for clustering model selection, using the three studied algorithms ($K$-means, Ward linkage and GMM). Only the extended versions of Stadion for $K$-means and GMM are included to speed up the analysis. Performance is evaluated using the ARI between the selected solution and the ground-truth clustering.

\subsection{Importance study with fANOVA}

Ideally, practitioners would like to know how hyperparameters affect performance, not just in the context of a single fixed instantiation of the remaining hyperparameters, but across all their instantiations. The fANOVA (functional ANalysis Of VAriance) framework for assessing hyperparameter importance introduced in \cite{hoos2014efficient} is based on efficient marginalization over dimensions using regression trees.
The importance of each hyperparameter is obtained by training a Random Forest model of 100 regression trees to predict the performance of Stadion given the set of hyperparameters. Then, the variance of the performance due to a given hyperparameter is decomposed by marginalizing out the effects of all other parameters. It also allows to assess interaction effects. Hence, the fANOVA framework provides insights on the overall importance of hyperparameters and their interactions.

The maximum amount of noise $\varepsilon_{\text{max}}$ and the fineness of the grid are not included in the study, because it is data-dependent and one can easily check if values are appropriate by looking at the paths. We study the following discrete hyperparameter space:
\begin{itemize}
    \item $D \in \lbrace 1, \ldots ,10 \rbrace $
    \item noise is uniform or Gaussian
    \item $\mathbf{\Omega} \in \lbrace 2,3,5,10, \lbrace 2, \ldots, 5 \rbrace, \lbrace 2, \ldots, 10 \rbrace, \lbrace 10 , \ldots ,20\rbrace,$\\ $\lbrace 2, \ldots 20 \rbrace \rbrace$
\end{itemize}
\begin{figure}[h]
    \centering
    \includegraphics[width=0.49\textwidth]{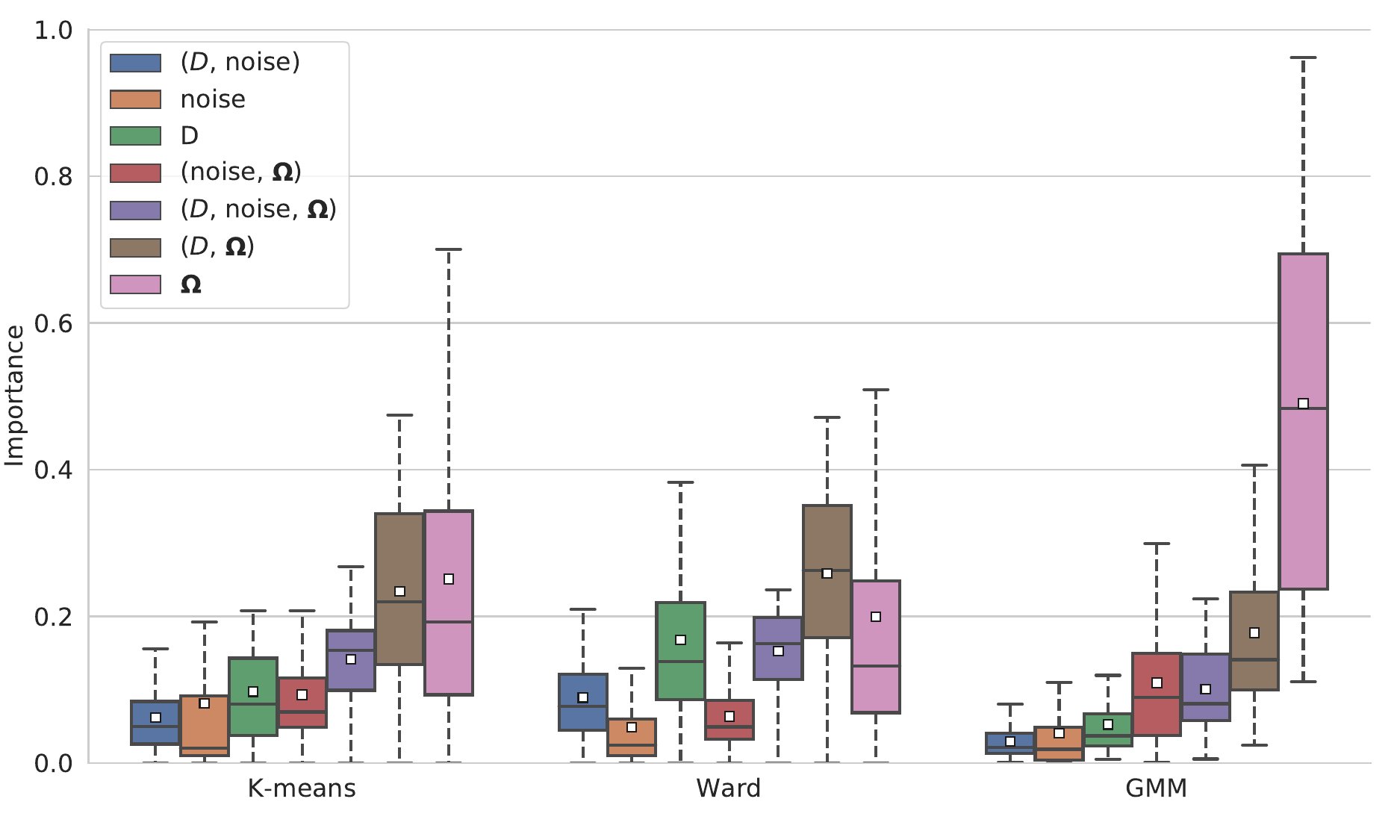}
    \includegraphics[width=0.49\textwidth]{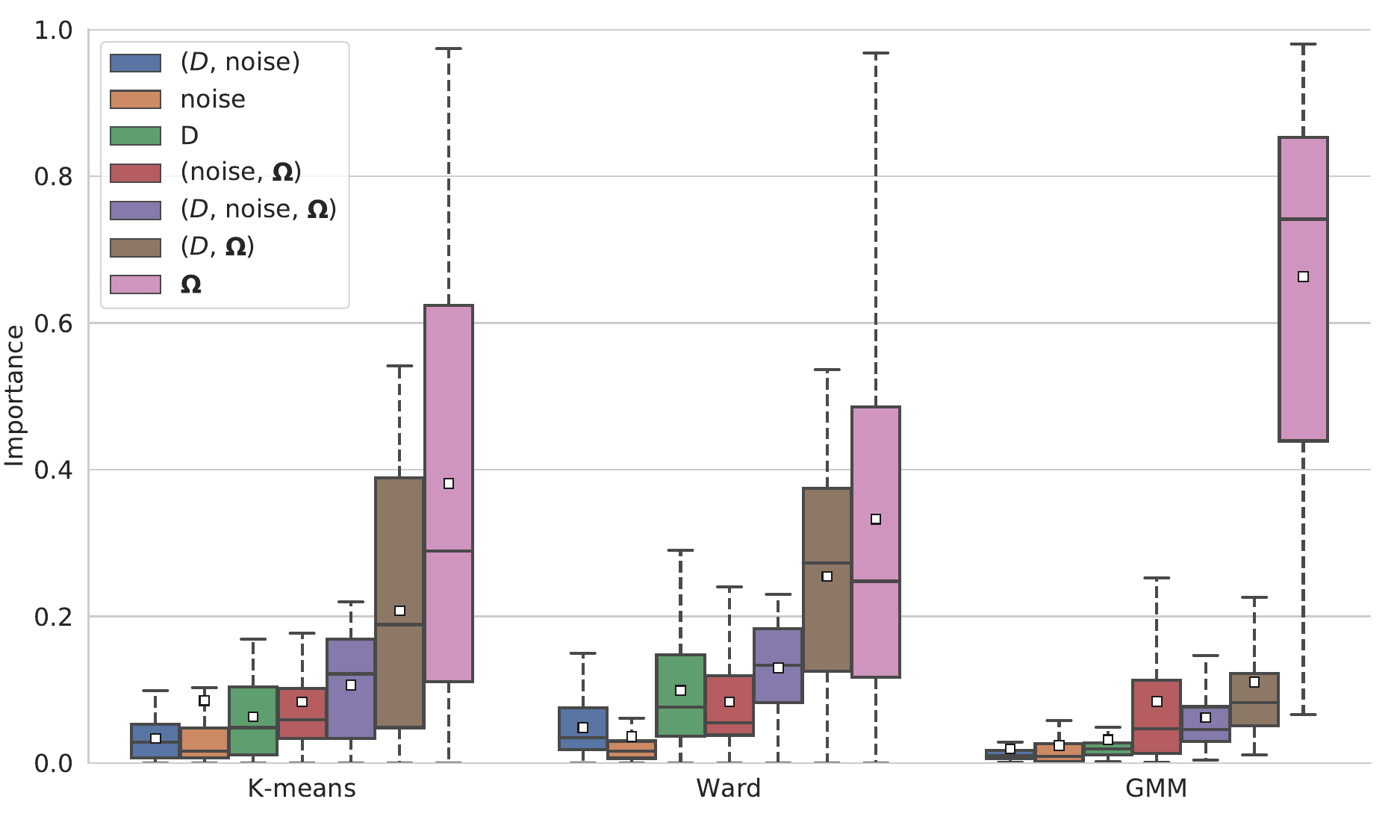}
    \caption{Box plots of the fANOVA importance of parameters and their interactions for Stadion-max (left) and mean (right) across 73 benchmark data sets, for three algorithms.}
    \label{fig:boxplot-importance}
\end{figure}

Figure~\ref{fig:boxplot-importance} shows the contributions of hyperparameters and their interactions to the variance of ARI performance, across 73 benchmark data sets. $\mathbf{\Omega}$ is by far the most important parameter, and this for two reasons.

First, the size of the data $N$ needed to obtain good estimations is relative to the number of clusters $K$. More precisely, in our theoretical setting we consider $K<<N$. Whenever $K'$ is large relative to $N_k$, which can happen in within-cluster stability for small clusters, jumping can occur. This implies that even in presence of sub-clusters, high values of $K'$ in $\mathbf{\Omega}$ will create instability and thus lead to low within-cluster stability. More precisely, if $K \geq K^{\star}$, then in general $\Omega$ will not affect within-cluster stability because it is already low. But whenever $K \leq K^{\star}$, within-cluster stability is more impacted by large values of $K'$ in $\Omega$, negatively affecting Stadion. However, impact on performance is very limited as shown in the following paragraph.

The second most important parameter is the interaction ($D$, $\mathbf{\Omega})$, for the same reason: large numbers of clusters make estimating the within-cluster stability more difficult, and thus a higher number of perturbations $D$ is needed to obtain a good approximation.


%
\subsection{Influence of $D$}

\begin{figure}[h]
    \centering
    \includegraphics[width=0.49\textwidth]{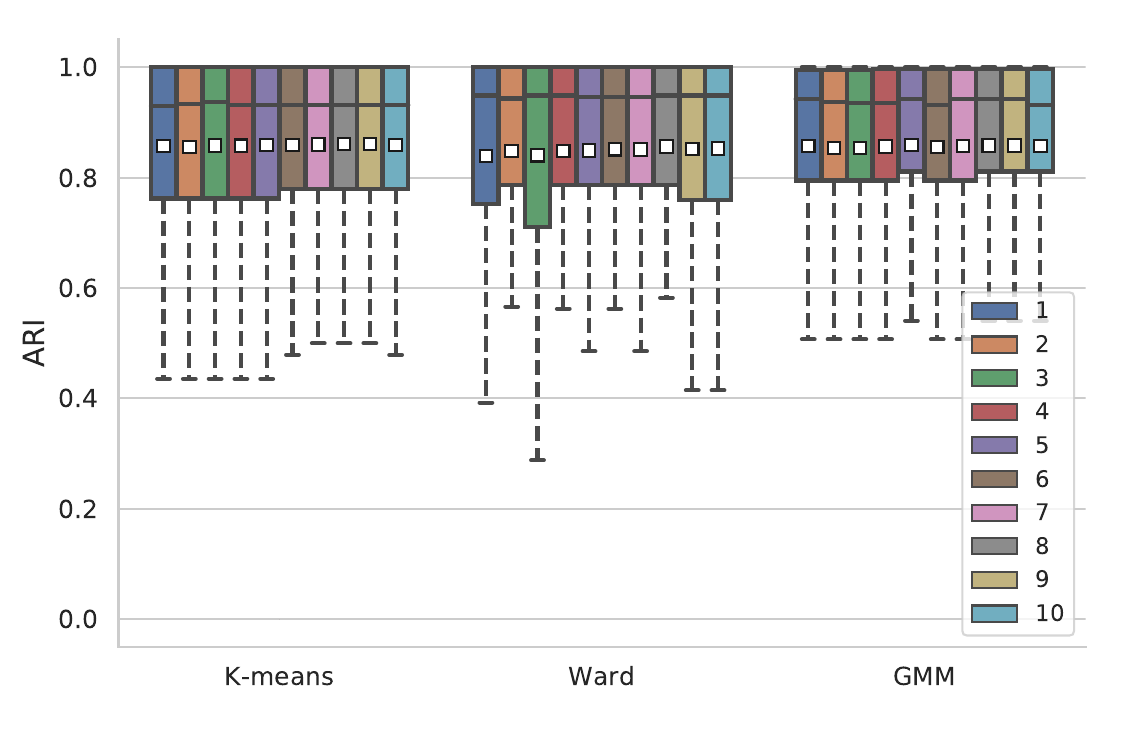}
    \includegraphics[width=0.49\textwidth]{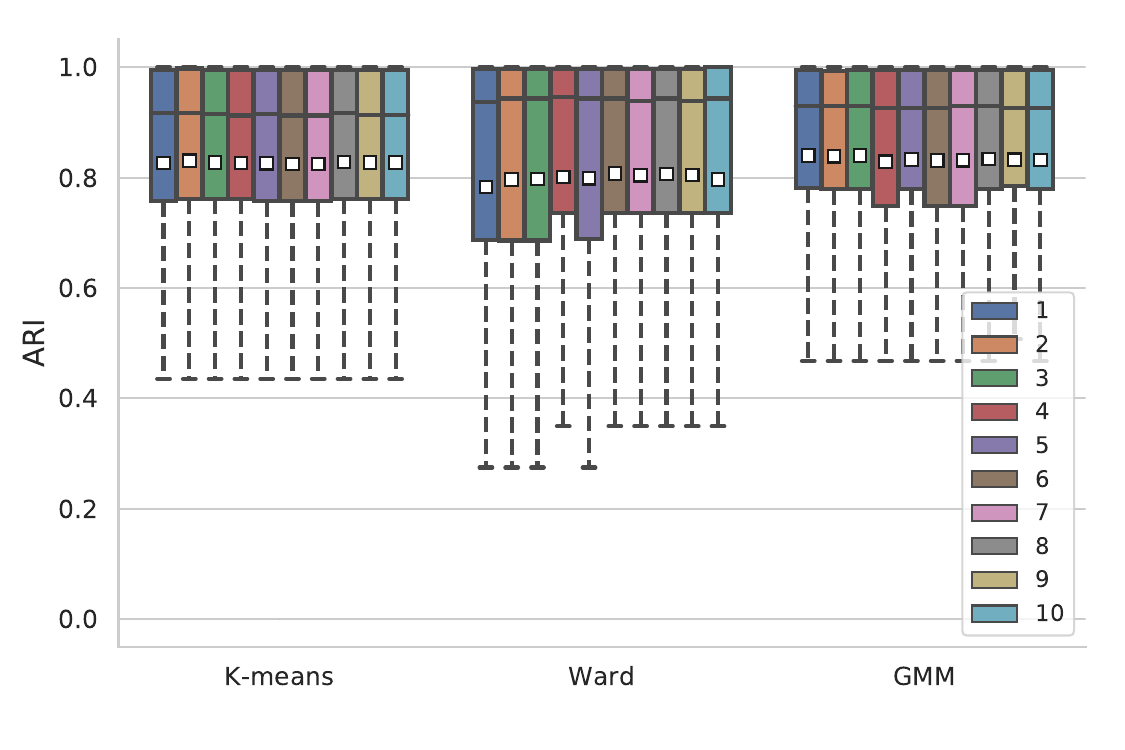}
    \caption{Box-plot of the ARI of partitions selected by Stadion-max (left) and mean (right)  across 73 data sets, for three algorithms and different values of $D$, the number of samples in the stability computation.}
    \label{fig:boxplot-D}
\end{figure}

\begin{figure}[h]
    \centering
    \includegraphics[width=0.45\textwidth]{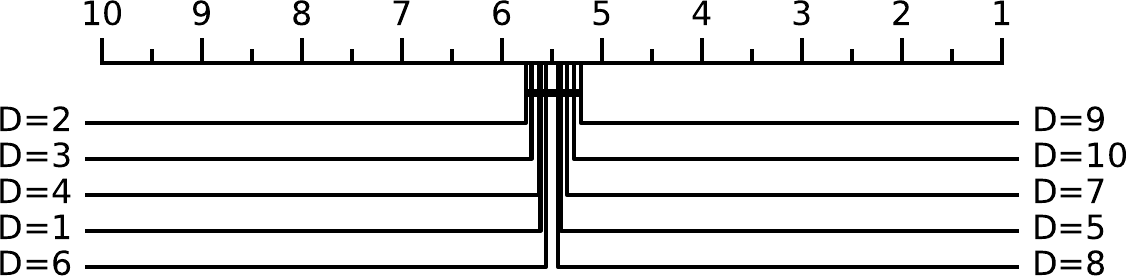}\;
    \includegraphics[width=0.45\textwidth]{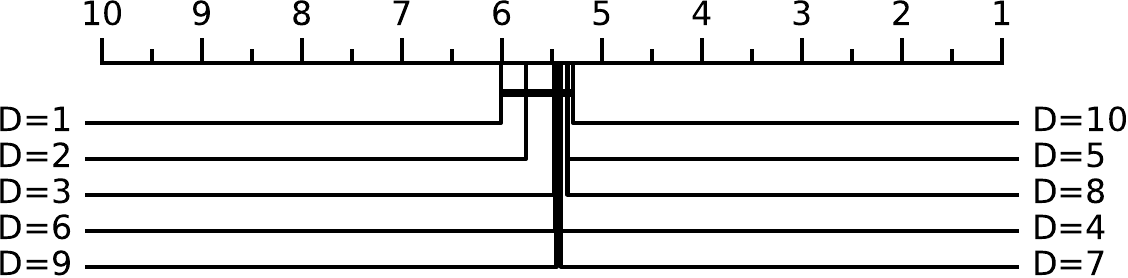}
    \caption{Critical difference diagrams after Wilcoxon-Holms test ($\alpha = 5\%$) on GMM performance, for Stadion-max with uniform (left) and Gaussian (right) noise, for different values of $D$, the number of perturbations in the stability computation.}
    \label{fig:cd-D}
\end{figure}

The $D$ hyperparameter defines the number of perturbed samples used in the stability computation. In our benchmark, we used $D = 10$. Surprisingly, a number of perturbed samples as low as $D = 1$ already gives a good estimate of the expectation and the performance only slightly increases with larger values of $D$. We perform an experiment by making $D$ vary from 1 to 10, keeping other hyperparameters fixed (uniform noise, $\mathbf{\Omega} = \{2, \ldots, 10\}$), for the three algorithms and both Stadion path aggregation strategies (max and mean), and measure performance in terms of ARI over 73 benchmark data sets. Results on Figure~\ref{fig:boxplot-D} show that low $D$ values have a higher variance and slightly lower performance.
To quantify further the influence of this parameter, we followed the recommendation in \cite{demvsar2006statistical} and used the Friedman test \cite{friedman1940comparison} for comparisons on multiple data sets, in order to test against the null hypothesis $\mathcal{H}_0$ stating that all parameters have equivalent performance. After rejecting $\mathcal{H}_0$, we performed the pairwise post-hoc analysis recommended by \cite{benavoli2016should} where the average rank comparison (e.g. Nemenyi test) is replaced by a Wilcoxon signed-rank test \cite{wilcoxon1947probability} at $\alpha = 5\%$ with a Holm-Bonferroni correction procedure to control the family-wise error rate (FWER) \cite{holm1979simple, garcia2008extension}. To visualize post-hoc test results, we use the critical difference (CD) diagram \cite{demvsar2006statistical}, where a thick horizontal line shows groups (cliques) of classifiers that are not significantly different in terms of performance. In all but one case, the Friedman test could not reject the null hypothesis. Only for the GMM algorithm and max aggregation, the null hypothesis was rejected, leading to the critical difference diagrams on Figure~\ref{fig:cd-D}.
The number of samples $D$ has a negligible impact on the performance of our method. We assume it is due to the fact that in our setting, instability is caused by jittering at cluster boundaries, which does not vary much from one perturbation to another with reasonable amounts of data.
On the contrary, sampling-based stability methods that rely on jumping require a much higher number of samples (for instance, \cite{Ben-Hur2002} use 100 samples and \cite{Lange2004} use 20 samples). As a conclusion, we recommend using $D \geq 5$, but if computation time is costly, $D=1$ can be used safely to cut down complexity.

\subsection{Influence of noise type}

\begin{figure}[h]
    \centering
    \includegraphics[width=0.49\textwidth]{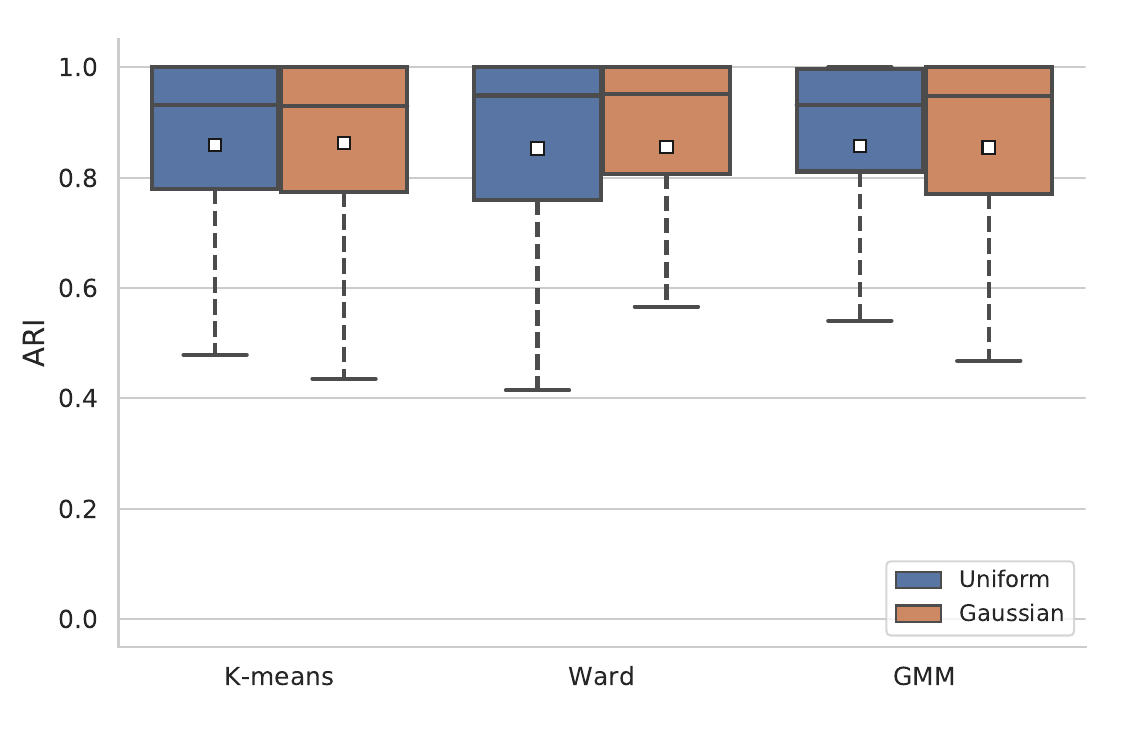}
    \includegraphics[width=0.49\textwidth]{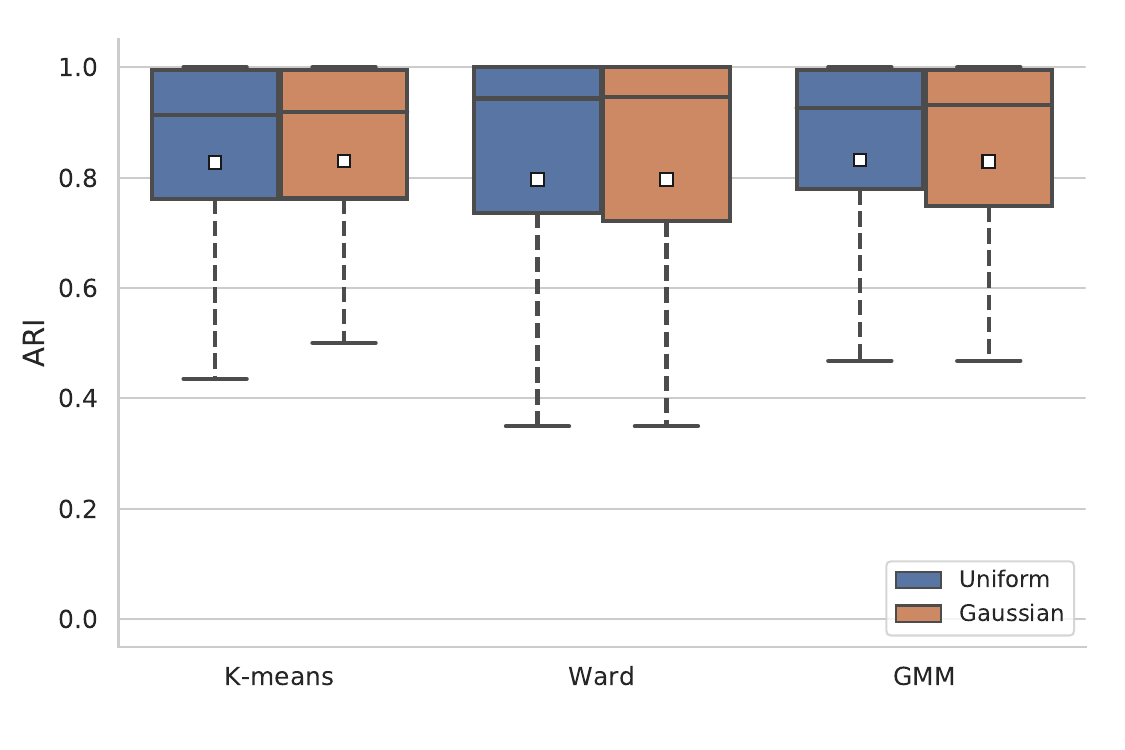}
    \caption{Box plots of the ARI of partitions selected by Stadion-max (left) and mean (right) across 73 data sets, for three algorithms, using uniform or Gaussian noise perturbation.}
    \label{fig:boxplot-noise}
\end{figure}

We experiment with two types of $\varepsilon$-additive noise perturbation: uniform noise and Gaussian noise. As previously, we report the distributions of performance in terms of ARI across 73 data sets for both noise types on Figure~\ref{fig:boxplot-noise} (with $D=10$ and $\mathbf{\Omega} = \{2, \ldots, 10\}$).
To assess the difference between both noise types, we perform the Wilcoxon signed-rank test on the performance results (at confidence level $\alpha = 5\%$). For every algorithm and Stadion path aggregation, the test did not reject the null hypothesis. Thus, either uniform or Gaussian noise can be used. 

\subsection{Influence of $\mathbf{\Omega}$}

\begin{figure}[h]
    \centering
    \subfigure[Stadion-max / uniform noise]{\includegraphics[width=0.49\textwidth]{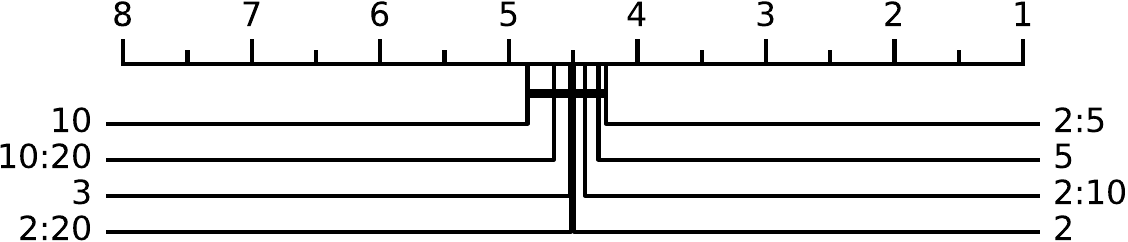}}
    \subfigure[Stadion-max / Gaussian noise]{\includegraphics[width=0.49\textwidth]{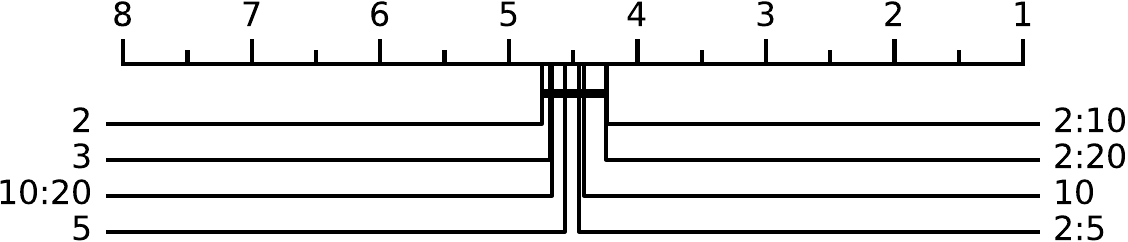}}
    \newline
    \subfigure[Stadion-mean / uniform noise]{Could not reject hypothesis $\mathcal{H}_0$.}
    \subfigure[Stadion-mean / Gaussian noise]{\includegraphics[width=0.49\textwidth]{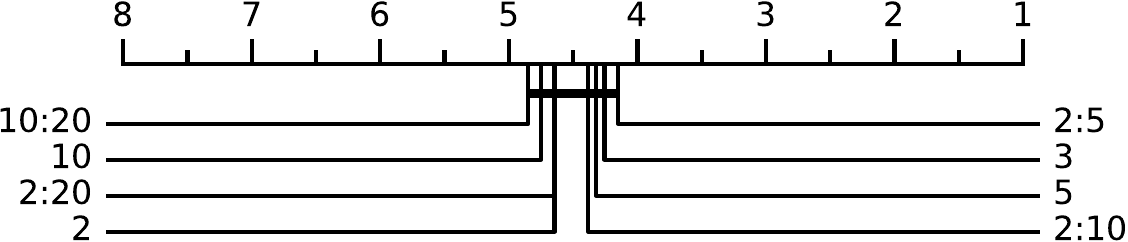}}
    \caption{Critical difference diagrams after Wilcoxon-Holms test on $K$-means performance, for different values of $\mathbf{\Omega}$.}
    \label{fig:cd-omega-km}
\end{figure}

\begin{figure}[h]
    \centering
    \subfigure[Stadion-max / uniform noise]{\includegraphics[width=0.49\textwidth]{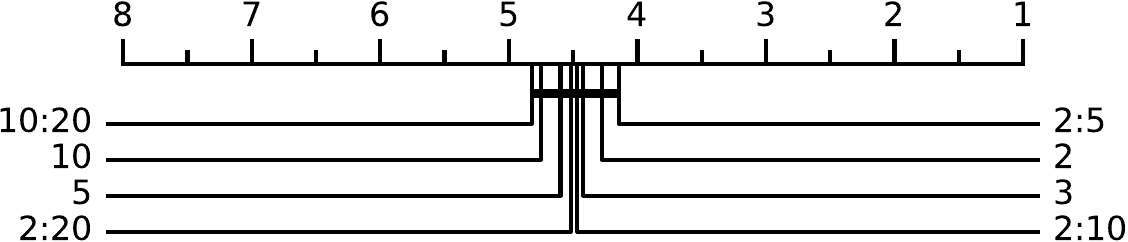}}
    \subfigure[Stadion-max / Gaussian noise]{Could not reject hypothesis $\mathcal{H}_0$.}
    \subfigure[Stadion-mean / uniform noise]{\includegraphics[width=0.49\textwidth]{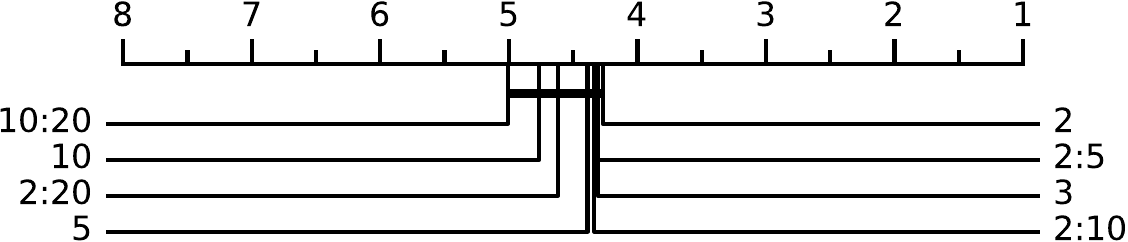}}
    \subfigure[Stadion-mean / Gaussian noise]{\includegraphics[width=0.49\textwidth]{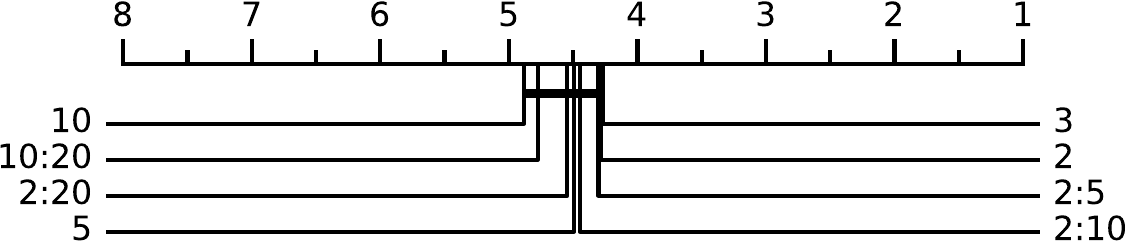}}
    \caption{Critical difference diagrams after Wilcoxon-Holms test on Ward performance, for different values of $\mathbf{\Omega}$.}
    \label{fig:cd-omega-ward}
\end{figure}

\begin{figure}[h]
    \centering
    \subfigure[Stadion-max / uniform noise]{\includegraphics[width=0.49\textwidth]{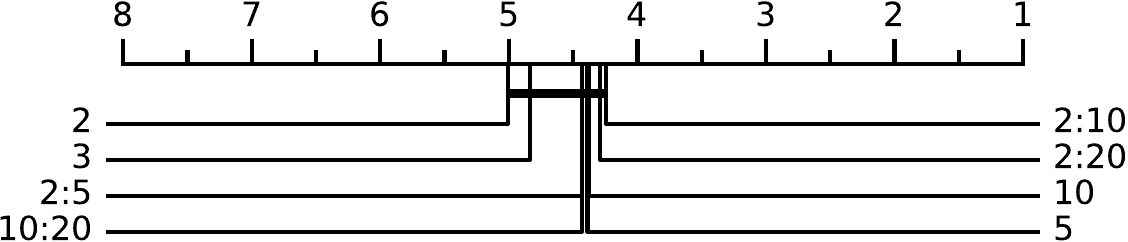}}
    \subfigure[Stadion-max / Gaussian noise]{\includegraphics[width=0.49\textwidth]{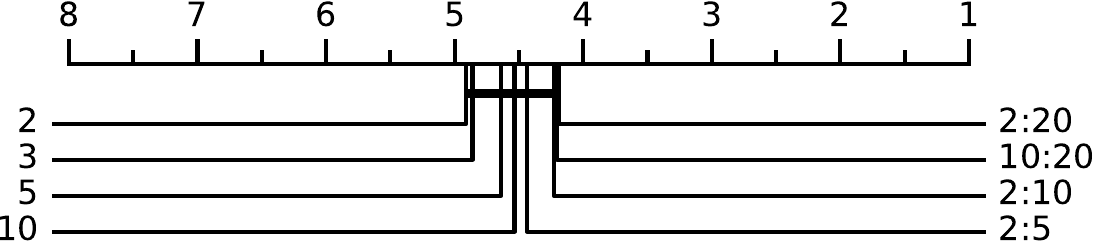}}
    \caption{Critical difference diagrams after Wilcoxon-Holms test on GMM performance, for different values of $\mathbf{\Omega}$. With Stadion-mean, the Friedman test could not reject $\mathcal{H}_0$.}
    \label{fig:cd-omega-gmm}
\end{figure}

The $\mathbf{\Omega}$ hyperparameter is a set defining the numbers of clusters used to cluster again each cluster of the original partition. We perform an experiment by varying $\mathbf{\Omega}$, keeping other hyperparameters fixed (uniform noise, $D=10$), for both Stadion path aggregation strategies (max and mean), and measure performance in terms of ARI over 73 benchmark data sets.
\begin{figure}[h]
    \centering
    \includegraphics[width=0.49\textwidth]{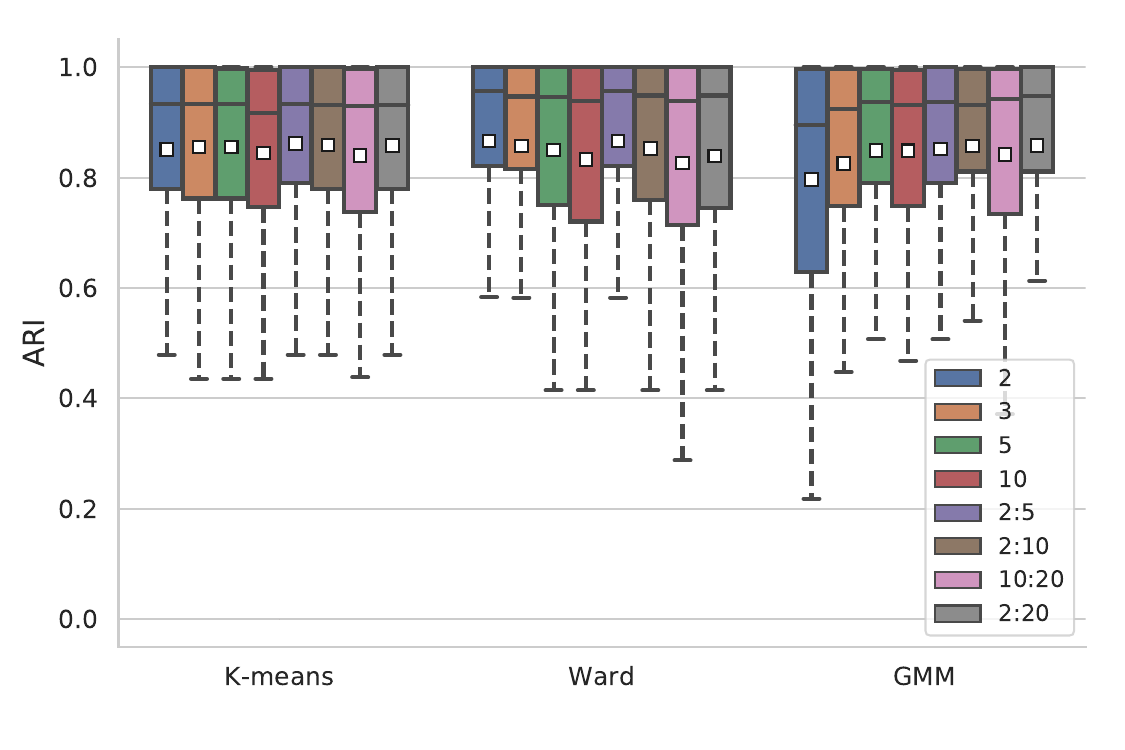}
    \includegraphics[width=0.49\textwidth]{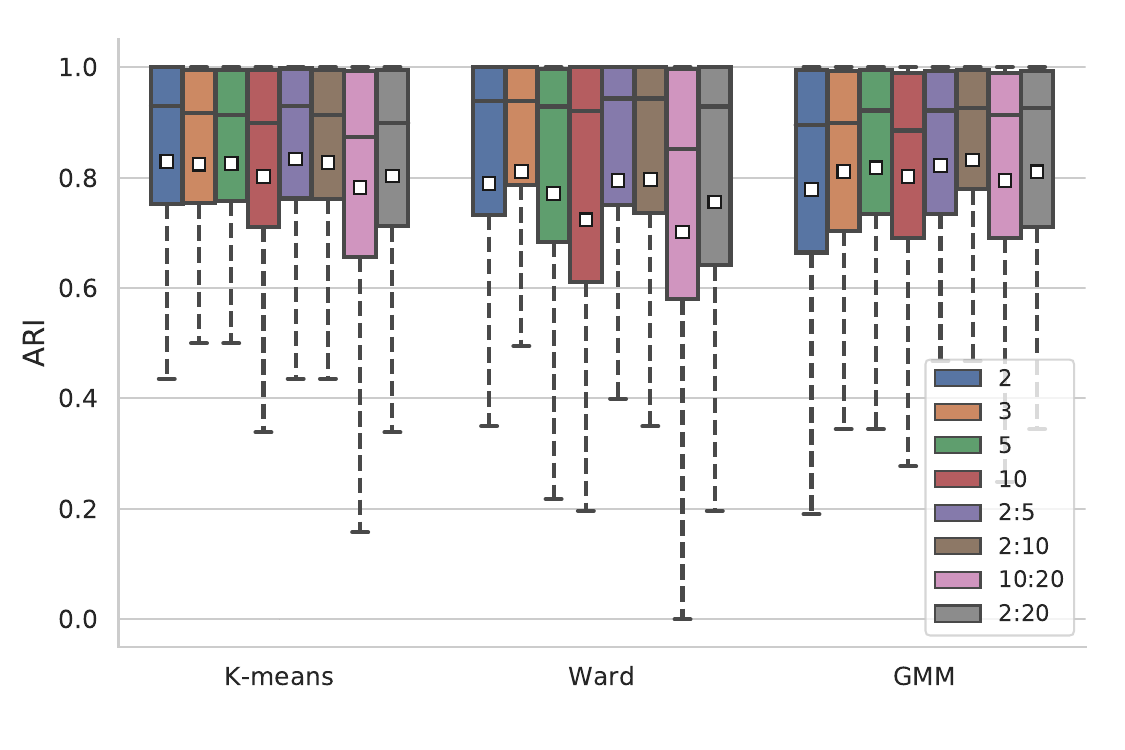}
    \caption{Box plots of ARI of partitions selected by Stadion-max (left) and mean (right) across 73 data sets, for three algorithms and different sets $\mathbf{\Omega}$ (numbers of clusters used in within-cluster stability).}
    \label{fig:boxplot-omega}
\end{figure}
Results in Figure~\ref{fig:boxplot-omega} demonstrate that $\mathbf{\Omega}$ does not have, in most cases, a big impact on the performance of Stadion. This does not contradict the results of the fANOVA. Indeed, $\mathbf{\Omega}$ has the largest variance contribution to the performance, but this variance remains small, and overall Stadion is robust for reasonable choices of the parameter, such as $\{2, \ldots, 5\}$ or $\{2, \ldots, 10\}$. Ward linkage is the most influenced by the choice of $\mathbf{\Omega}$. We guess the main reason is that agglomerative clustering algorithms are not robust to noise \cite{balcan2016clustering}. Critical difference diagrams after Wilcoxon-Holms test on performance are given in Figures~(\ref{fig:cd-omega-km}, \ref{fig:cd-omega-ward}, \ref{fig:cd-omega-gmm}). None of them showed significant differences, indicating that there is not enough data to conclude. However, small values in $\mathbf{\Omega}$ seem to perform better, which confirms the previous claim that large values of $K'$ in $\mathbf{\Omega}$ negatively impact performance. In particular, the range $\{2, \ldots, 10\}$ used in our benchmark performs well across all algorithms.

%% file: supplementary/code.tex
\section{Complexity analysis}\label{appendix:code}

This section provides a time complexity analysis of the Stadion computation.
Let $A(K,N)$ be the time complexity of the algorithm with parameter $K$ and data of size $N$, assuming the data dimension is fixed. In addition, let $S(K,N)$ be the complexity of the similarity measure $s$, $D$ the number of perturbations and $M$ the number of noise levels.

\paragraph{Between-cluster stability} The complexity is $\mathcal{O}\left((A(K,N) + S(K,N))  D  M\right)$.

\paragraph{Within-cluster stability} For a given parameter $K$ and a set of parameters $\mathbf{\Omega} = \{ 2, \dots, K'\}$, the amount of operations is $\sum_{k=1}^K \sum_{k'=2}^{K'} (A(k',N_k) + S(k',N_k))  D  M$, which can be bounded by $\mathcal{O}\left(K  K'  (A(K',N) + S(K',N))  D  M\right)$.\\

In the case of $K$-means, we have $A(K,N) = \mathcal{O}(K N T I)$, where $T$ is the number of iterations until convergence, and $I$ the number of initialization runs. Then, ARI is linear: $S(K, N) = \mathcal{O}(N)$. Overall, we obtain a complexity for Stadion with $K$-means and ARI equal to $\mathcal{O}(K K'^2 N T I D M)$.

The influence studies showed that $\mathbf{\Omega}$ can be set to a small range, e.g. $\{2, \ldots, 5\}$ or $\{2, \ldots, 10\}$, and that $D$ can be kept very low. In addition, the number of noise levels $M$ is fixed. Thus, complexity in $K'$, $D$ and $M$ is manageabl, and complexity of Stadion is mainly driven by the complexity of the algorithm itself.

In regard, internal indices relying on between-cluster and within-cluster distances have a complexity of $\mathcal{O}(N^2)$ with pairwise distance or $\mathcal{O}(KN)$ with centroid distance. Thus, the cost of having to run the algorithm several times may be smaller than a quadratic index if the algorithm is linear in $N$.


%% file: supplementary/experiment.tex
\section{Experimental setting}\label{appendix:experiment}

In this section, we provide additional information on our experimental setting.

\subsection{Algorithm initialization}

\begin{itemize}
    \item $K$-means: initialization is achieved using $K$-means++ \cite{Arthur2007} and keeping the best of 35 runs (w.r.t. the cost function).
    \item Ward linkage: agglomerative clustering is deterministic, no initialization strategy is needed.
    \item Gaussian Mixture Model: 
    Two initialization schemes were considered. The first one uses $K$-means to initialize the EM algorithm. The second one uses the approach discussed in \cite{scrucca2015improved} based on a scaled SVD and agglomerative hierarchical clustering at the initialization step. However, there was no noticeable difference between the two initializations.
\end{itemize}

\subsection{Preprocessing}

Every data set was scaled to zero mean and unit variance on each dimension.

\subsection{Choice of the external performance metric}

Note that performance is evaluated using the external index ARI w.r.t. the ground-truth cluster assignments, while Stadion also uses $s$ = ARI as its similarity measure to estimate stability. It would be reasonable to expect this situation to introduce some kind of bias. However, we also measured the performance using many other external metrics (such as NMI), and almost no change in the final ranking was observed. Thus, results are not biased in favor of Stadion, and we kept ARI as a standard, adjusted, performance metric \cite{vinh2010information}.

\subsection{List of data sets}

A complete list of the 73 data sets used in the benchmark is provided below, indicating the number of samples ($N$), dimension ($p$), ground-truth number of clusters ($K^{\star}$) and reference. The data sets are easily downloaded from the archives \cite{deric} and \cite{gago}. The data sets without references are original and have been created for this work, in order to provide challenging model selection tasks. They will be available in the companion repository of this paper.

\begin{table}[h]
\caption{List of the benchmark data sets.}
\begingroup
\setlength{\tabcolsep}{3.8pt} 
\begin{center}
\begin{tiny}
\begin{tabular}{lrrrr}
\toprule
Dataset & $N$ & $p$ & $K^{\star}$ & reference \\
\midrule
2d-10c & 2990 & 2 & 9 & \cite{clustergenerators} \\
2d-3c-no123 & 715 & 2 & 3 & \cite{clustergenerators} \\
2d-4c & 1261 & 2 & 4 & \cite{clustergenerators} \\
2d-4c-no4 & 863 & 2 & 4 & \cite{clustergenerators} \\
2d-4c-no9 & 876 & 2 & 4 & \cite{clustergenerators} \\
\addlinespace
3clusters\_elephant & 700 & 2 & 3\\
4clusters\_corner & 1575 & 2 & 4\\
4clusters\_twins & 612 & 2 & 4\\
5clusters\_stars & 1050 & 2 & 5\\
A1 & 3000 & 2 & 20 & \cite{sipu} \\
\addlinespace
A2 & 5250 & 2 & 35 & \cite{sipu} \\
curves1 & 1000 & 2 & 2 & \cite{deric} \\
D31 & 3100 & 2 & 31 & \cite{veenman2002} \\
diamond9 & 3000 & 2 & 9 & \cite{salvador}  \\
dim032 & 1024 & 32 & 16 & \cite{sipu}  \\
\addlinespace
dim064 & 1024 & 64 & 16 & \cite{sipu}  \\
dim1024 & 1024 & 1024 & 16 & \cite{sipu}  \\
dim128 & 1024 & 128 & 16 & \cite{sipu}  \\
dim256 & 1024 & 256 & 16 & \cite{sipu}  \\
dim512 & 1024 & 512 & 16 & \cite{sipu}  \\
\addlinespace
DS-577 & 577 & 2 & 3 & \cite{su2005}\\
DS-850 & 850 & 2 & 5 & \cite{su2005}\\
ds4c2sc8 & 485 & 2 & 8 & \cite{faceli2010} \\
elliptical\_10\_2 & 500 & 2 & 10 & \cite{bandyopadhyay2007} \\
elly-2d10c13s & 2796 & 2 & 10 & \cite{clustergenerators} \\
\addlinespace
engytime & 4096 & 2 & 2  & FCPS \cite{ultsch2005} \\
exemples1\_3g & 525 & 2 & 3\\
exemples10\_WellS\_3g & 975 & 2 & 3\\
exemples2\_5g & 1375 & 2 & 5\\
exemples3\_Uvar\_4g & 1000 & 2 & 4\\
\addlinespace
exemples4\_overlap\_3g & 1050 & 2 & 3\\
exemples5\_overlap2\_3g & 1550 & 2 & 3\\
exemples6\_quicunx\_4g & 2250 & 2 & 4\\
exemples7\_elbow\_3g & 788 & 2 & 3\\
exemples8\_Overlap\_Uvar\_5g & 2208 & 2 & 6\\
\addlinespace
exemples9\_YoD\_6g & 2208 & 2 & 6\\
fourty & 1000 & 2 & 40 & \cite{deric} \\
g2-16 & 2048 & 16 & 2 & \cite{sipu} \\
g2-2 & 2048 & 2 & 2 & \cite{sipu} \\
g2-64 & 2048 & 64 & 2 & \cite{sipu} \\
\addlinespace
hepta & 212 & 3 & 7 & FCPS \cite{ultsch2005} \\
long1 & 1000 & 2 & 2  & \cite{handl2004} \\
long2 & 1000 & 2 & 2  & \cite{handl2004} \\
long3 & 1000 & 2 & 2  & \cite{handl2004} \\
longsquare & 900 & 2 & 6  & \cite{handl2004} \\
\addlinespace
R15 & 600 & 2 & 15 & \cite{veenman2002} \\
s-set1 & 5000 & 2 & 15  & \cite{sipu} \\
s-set2 & 5000 & 2 & 15  & \cite{sipu} \\
s-set3 & 5000 & 2 & 15  & \cite{sipu} \\
s-set4 & 5000 & 2 & 15  & \cite{sipu} \\
\addlinespace
sizes1 & 1000 & 2 & 4  & \cite{handl2004} \\
sizes2 & 1000 & 2 & 4  & \cite{handl2004} \\
sizes3 & 1000 & 2 & 4  & \cite{handl2004} \\
sizes4 & 1000 & 2 & 4  & \cite{handl2004} \\
sizes5 & 1000 & 2 & 4  & \cite{handl2004} \\
\addlinespace
spherical\_4\_3 & 400 & 3 & 4 & \cite{bandyopadhyay2007} \\
spherical\_5\_2 & 250 & 2 & 5 & \cite{bandyopadhyay2007} \\
spherical\_6\_2 & 300 & 2 & 6 & \cite{bandyopadhyay2007} \\
square1 & 1000 & 2 & 4  & \cite{handl2004} \\
square2 & 1000 & 2 & 4  & \cite{handl2004} \\
\addlinespace
square3 & 1000 & 2 & 4  & \cite{handl2004} \\
square4 & 1000 & 2 & 4  & \cite{handl2004} \\
square5 & 1000 & 2 & 4  & \cite{handl2004} \\
st900 & 900 & 2 & 9  & \cite{bandyopadhyay2007} \\
tetra & 400 & 3 & 4  & FCPS \cite{ultsch2005} \\
\addlinespace
triangle1 & 1000 & 2 & 4 & \cite{handl2004} \\
triangle2 & 1000 & 2 & 4 & \cite{handl2004} \\
twenty & 1000 & 2 & 20  & \cite{deric} \\
twodiamonds & 800 & 2 & 2  & FCPS \cite{ultsch2005} \\
wingnut & 1016 & 2 & 2  & FCPS \cite{ultsch2005} \\
\addlinespace
xclara & 3000 & 2 & 3  & \cite{xclara} \\
zelnik2 & 303 & 2 & 3 & \cite{zelnik2004} \\
zelnik4 & 622 & 2 & 5 & \cite{zelnik2004} \\
\bottomrule
\end{tabular}
\end{tiny}
\end{center}
\endgroup
\end{table}